\documentclass[twocolumn]{svjour3}
\smartqed  
\usepackage{booktabs}
\usepackage{graphicx}
\usepackage{amsmath}
\usepackage{amssymb}
\usepackage{tabularx}
\usepackage{multirow}
\usepackage{algorithm}
\usepackage{algorithmic}
\usepackage{stackengine}
\usepackage{xcolor}
\usepackage{adjustbox}
\usepackage{colortbl}
\usepackage{ragged2e}
\usepackage{rotating,soul}
\usepackage[caption=false,font=footnotesize]{subfig}
\usepackage{tipa}
\usepackage{ulem}
\usepackage{color}
\usepackage{booktabs}
\usepackage{float}
\usepackage{bm}
\usepackage{bbding}
\usepackage{verbatim}
\usepackage{makecell}
\usepackage{overpic}
\usepackage{setspace}

\usepackage[numbers,sort]{natbib}
\bibpunct[,]{(}{)}{;}{a}{}{,}

\usepackage{pifont}%

\definecolor{mygray}{gray}{.93}
\definecolor{mygray1}{gray}{.99}

\definecolor{mygreen}{RGB}{60,200,20}
\definecolor{mypurple}{RGB}{170,80,230}

\definecolor{wpy_1}{rgb}{1,0,0}

\newcommand{\myPara}[1]{\noindent\textbf{#1}}

\def\ie{\textit{i.e.}}
\def\eg{\textit{e.g.}}

\definecolor{linkcolor}{RGB}{255,0,0}
\definecolor{urlcolor}{RGB}{255,105,180}
\definecolor{citecolor}{RGB}{0, 80, 200}
\definecolor{myorange}{RGB}{0,0,0}
\usepackage{hyperref}
\hypersetup{colorlinks=true,linkcolor=linkcolor,urlcolor=urlcolor,citecolor=citecolor}

\journalname{International Journal of Computer Vision}

\begin{document}\sloppy

\title{Background Activation Suppression for Weakly Supervised Object Localization and Semantic Segmentation}

\titlerunning{Background Activation Suppression for Weakly Supervised Object Localization and Semantic Segmentation}        
	\author{
		Wei Zhai$^{1\dag}$\and
		Pingyu Wu$^{1\dag}$\and
		Kai Zhu$^{1}$\and
		Yang Cao$^{1,2}$\and
		Feng Wu$^{1,2}$\and \\
		Zheng-Jun Zha$^{1}$
	}
	
	\authorrunning{Zhai et al.} 

	\institute{
	Wei Zhai (wzhai056@ustc.edu.cn) \\
	Pingyu Wu (wpy364755620@mail.ustc.edu.cn) \\
	Kai Zhu (zkzy@mail.ustc.edu.cn) \\
	Yang Cao (forrest@ustc.edu.cn) \\
	Feng Wu (fengwu@ustc.edu.cn) \\
	Zheng-Jun Zha (zhazj@ustc.edu.cn) \\
	$^{\textbf{1}}$University of Science and Technology of China, Hefei, China \\
	$^{\textbf{2}}$Institute of Artificial Intelligence, Hefei Comprehensive National Science Center, China \\
	$^{\dag}$Wei Zhai and Pingyu Wu contributed equally to this work.\\
    }

\date{Received: date / Accepted: date}

\maketitle

\begin{abstract}

Weakly supervised object localization and semantic segmentation aim to localize objects using only image-level labels. Recently, a new paradigm has emerged by generating a foreground prediction map (FPM) to achieve pixel-level localization. While existing FPM-based methods use cross-entropy to evaluate the foreground prediction map and to guide the learning of the generator, this paper presents two astonishing experimental observations on the object localization learning process: For a trained network, as the foreground mask expands,
1) the cross-entropy converges to zero when the foreground mask covers only part of the object region.
2) The activation value continuously increases until the foreground mask expands to the object boundary. Therefore, to achieve a more effective localization performance, we argue for the usage of activation value to learn more object regions. In this paper, we propose a Background Activation Suppression (BAS) method. Specifically, an Activation Map Constraint (AMC) module is designed to facilitate the learning of generator by suppressing the background activation value. Meanwhile, by using foreground region guidance and area constraint, BAS can learn the whole region of the object. In the inference phase, we consider the prediction maps of different categories together to obtain the final localization results. Extensive experiments show that BAS achieves significant and consistent improvement over the baseline methods on the \texttt{CUB-200-2011} and \texttt{ILSVRC} datasets. In addition, our method also achieves state-of-the-art weakly supervised semantic segmentation performance on the \texttt{PASCAL VOC 2012} and \texttt{MS COCO 2014} datasets. Code and models are available at \href{https://github.com/wpy1999/BAS-Extension}{\color{magenta}github.com/wpy1999/BAS-Extension}.


\end{abstract}

\section{Introduction}

Weakly supervised object localization (WSOL) aims to identify the object's localization in a scene, where only image-level labels instead of bounding box annotations are available during training. Due to the reduction in the cost of manual labeling, and the potential to use the vast weakly-annotated images on many public datasets and the Web, WSOL is gaining more and more attention in the research community~\citep{zhai2022one,luo2022learning,zhang2021weakly,yang2023grounding,zhai2022exploring,dong2023exploring,gou2022driver,zhou2023pit}. Moreover, it can serve various downstream tasks, such as weakly supervised object detection (WSOD)~\citep{song2021weakly,zhang2020discriminant,zhang2019leveraging} and weakly supervised semantic segmentation (WSSS)~\citep{ru2022weakly,chan2021comprehensive,pan2022learning}.

\begin{figure}[t]
	\centering
		\begin{overpic}[width=0.99\linewidth]{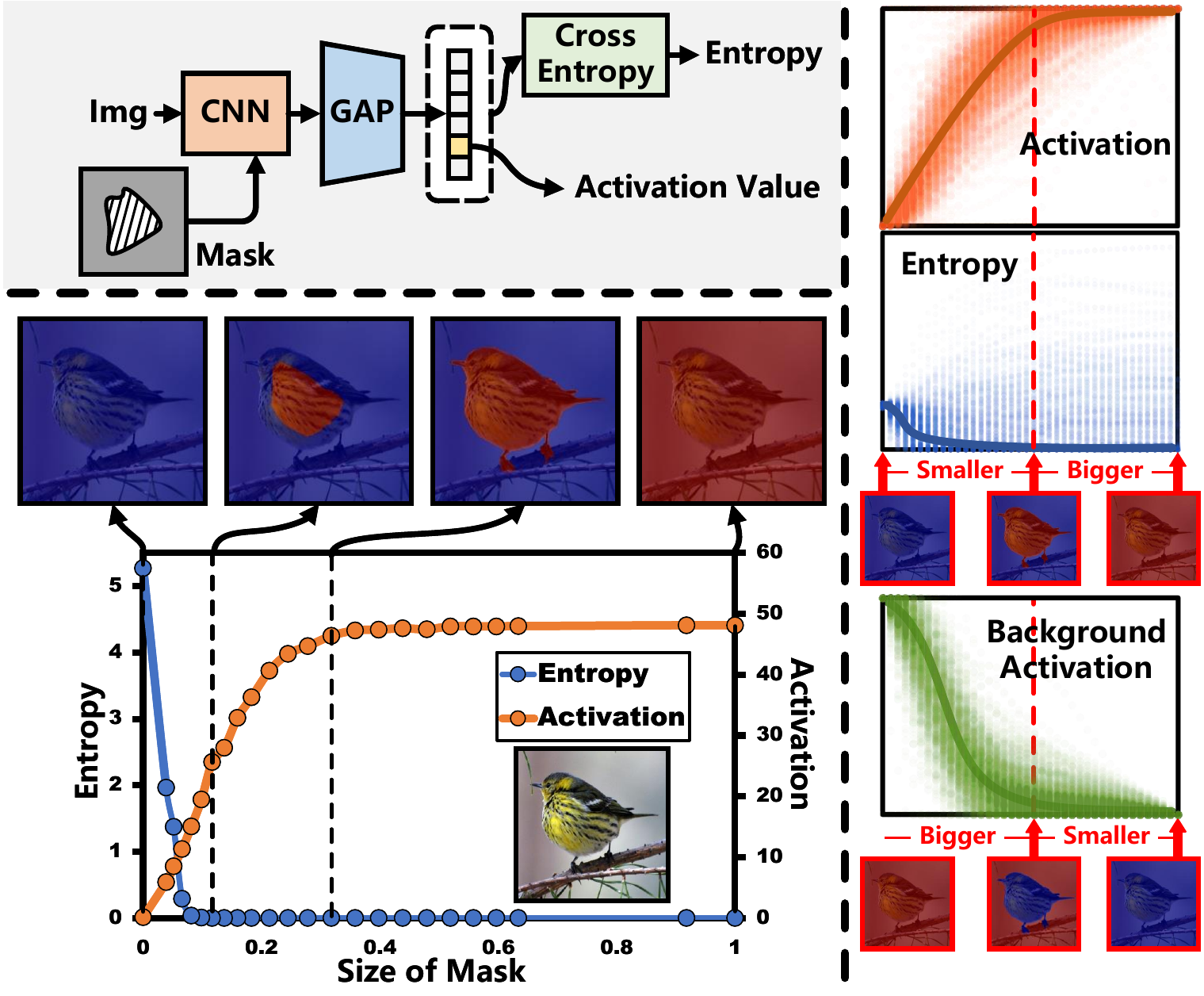}
		    \put(0.3, 70){\small\textbf{(A)}}
		    \put(0.3, 12){\small\textbf{(B)}}
		    \put(82.5, 0.3){\small\textbf{(C)}}
	\end{overpic}
	\caption{\textbf{(A)} Experimental procedure and related definitions. \textbf{(B)} The entropy value of CE loss $w.r.t$ foreground mask and foreground activation value $w.r.t$ foreground mask.  \textbf{(C)} The results with statistical significance. Implementation details of the experiment and further results are available in Section~\ref{sec:exploratory}.}
	\label{fig1}
\end{figure}

This paper aims to propose an effective approach for WSOL and its downstream task WSSS, since WSOL and WSSS tasks have similarities in that they both use image-level labels as supervision and need to obtain a high-quality pixel-level localization map from the classification network. Actually, WSSS task can be implemented by directly training a fully supervised semantic segmentation with the localization maps generated from WSOL as pseudo labels. Due to these reasons, they face a similar challenge of establishing supervision between image-level labels and pixel-level localization maps in an effective way.

Previously, most WSOL and WSSS methods utilize Class Activation Map (CAM)~\citep{zhou2016learning} to extract localization map from classifier. While CAM can localize approximate object regions, it always prefers to capture the most discriminative regions rather than the overall area of the object, resulting in limited localization performance. Therefore, numerous CAM-based approaches have been proposed to alleviate this problem. Adversarial erasing
methods~\citep{singh2017hide,zhang2018adversarial,choe2019attention,mai2020erasing,yun2019cutmix} erase the most discriminative regions during the training, forcing the network to learn more object features that facilitate complete localization. Some methods~\citep{zhang2020inter,pan2021unveiling,lee2022anti} improve the localization performance of CAM by establishing pixel-level spatial and semantic correlation. Additionally, some other methods~\citep{zhang2018self,wei2021shallow,kolesnikov2016seed} suggest using the thought of region growing to spread confidence regions and mine relevant features.

Although CAM-based method can conveniently extract the localization map from the classifier, this approach will lead to limitations and conflicts in optimization since the classifier needs to implement both localization and classification tasks. Very recently, a CAM-independent paradigm~\citep{meng2021foreground,xie2021online} is devised for WSOL to achieve localization with a foreground prediction map (FPM) obtained directly through a generator, which allows the two tasks to be accomplished separately in a unified model. Typically, ORNet~\citep{xie2021online} is a two-stage approach, which first trains a classification network as an evaluator, and then utilizes CE loss to guide the learning of generator by masking the original image with a foreground prediction map. Orthogonally, the foreground prediction map in the FAM~\citep{meng2021foreground} is split into several parts and separately masks high-level feature maps to achieve learning of different regions through CE loss. Despite FPM-based methods achieving promising performance, they still suffer from incomplete object localization.

To better understand FPM-based methods, we focus on exploring the entropy value of CE loss (entropy) with respect to ($w.r.t$) foreground mask. As shown in Fig. \ref{fig1} (A), by changing the area of the foreground mask and masking the feature map, the relationship between the entropy and foreground mask area is plotted in Fig. \ref{fig1} (B). An important phenomenon can be observed that there is a ``\textbf{mismatch}'' between entropy and ground-truth mask, \ie, entropy is already close to zero when foreground mask retains only part of the object region, which indicates that entropy cannot force the foreground map to learn the complete object area. The reason is that the exponential form of softmax amplifies the discrepancy in activation values and drives premature convergence of entropy. To find a better factor to facilitate localization learning, we further explore the activation value (before softmax calculation) $w.r.t$ foreground mask. As shown in Fig. \ref{fig1} (B), there is a higher ``\textbf{correlation}'' between activation value and foreground mask, \ie, activation value tends to saturate when the mask expands to the object boundary. This suggests that better localization ability can be learned by optimizing activation value. Fig. \ref{fig1} (C) also confirms the generality of these phenomena in a statistical sense.

Based on the inspiration of the above exploratory analysis, a straightforward manner to obtain a complete foreground prediction map is to maximize the activation value. However, considering that the minimization optimization problem is more conducive to the stability of training and loss convergence than the maximization optimization problem, this paper proposes a novel way to learn a background prediction map by minimizing background activation value, and further obtain the accurate foreground prediction map by inversion. Actually, the statistics on background activation values in Fig.~\ref{fig1} (C) show ``\textbf{symmetry}'' with the statistics on activation values, both converging at the ground-truth mask area, which further supports the feasibility of background activation value suppression.

In this paper, we propose a simple but effective \textbf{B}ackground \textbf{A}ctivation \textbf{S}uppression (\textbf{BAS}) method. As shown in Fig.~\ref{fig:network}, our method includes three modules: an extractor, a generator, and an \textbf{A}ctivation \textbf{M}ap \textbf{C}onstraint (\textbf{AMC}) module. First, an extractor is used to extract the image features for subsequent localization and classification. The generator aims to generate a class-specific foreground prediction map for localization. Then the coupled background prediction map is obtained by inverting the foreground prediction map and fed into AMC together for localization training. The AMC is supervised by four kinds of losses, which are background activation suppression loss, area constraint loss, foreground region guidance loss, and classification loss. The most important one is background activation suppression loss, which is devised to promote the learning of generator by minimizing the ratio of background activation value and overall activation value (the activation value generated by the entire image). In the inference phase, the Top-$k$ prediction maps are selected based on the predicted category probabilities and their average prediction map is adopted as the final localization result. The main contributions of this paper can be summarized as follows:

\begin{itemize}
	\item [\textbf{1)}] This paper identifies that the essential reason why minimizing CE loss facilitates the generation of foreground map is that it indirectly increases the foreground activation value, and accordingly proposes to promote the generation of foreground prediction map by suppressing background activation value.

	\item [\textbf{2)}] This paper proposes a simple but effective Background Activation Suppression (BAS) approach to facilitate the generation of foreground map by an Activation Map Constraint (AMC) in a weakly supervised manner, which is composed of four losses including background activation suppression loss and together contribute to the generation of the foreground prediction map for localization.

    \item [\textbf{3)}] Extensive experiments on \texttt{CUB-200-2011}~\citep{welinder2011caltech} and \texttt{ILSVRC}~\citep{russakovsky2015imagenet} benchmarks demonstrate that our method achieves consistent and significant improvement in terms of GT-known/Top-1/Top-5 Loc. In addition, the proposed BAS approach can be extended to Weakly Supervised Semantic Segmentation (WSSS) task, which also achieves new state-of-the-art results on \texttt{PASCAL VOC 2012}~\citep{everingham2010pascal} and \texttt{MS COCO 2014}~\citep{lin2014microsoft} datasets.

\end{itemize}

This paper builds upon our conference version~\citep{wu2022background}, which has been extended in four distinct aspects. \textbf{1)} We explain the advantages of Background Activation Suppression and its generalizability (on more complex datasets) in more detail and comprehensively (in a statistical sense), see Fig. \ref{fig:basexp} and Section \ref{sec:exploratory}. \textbf{2)} To alleviate the problem of inadequate convergence of BAS loss (Fig. \ref{fig:bas_loss}), we focus on the location of the \texttt{ReLU} function, which is closely related to the activation value, and further improve the previous BAS after exploration, see Fig. \ref{fig:relu} and Section \ref{AMC_section}. \textbf{3)} To verify the extensibility of the BAS approach, we develop a Weakly Supervised Semantic Segmentation (WSSS) framework with proposed BAS in Section \ref{sec:WSSS_experiment}. The framework aims to enhance the quality of the seed generation process in the popular WSSS framework through BAS, resulting in better performance on WSSS task, as shown in Table \ref{table:voc_acc_train}, Table \ref{table:voc_acc}, and Table \ref{table:coco_val}. \textbf{4)} To exploit the advantages of BAS on WSSS in obtaining localization maps through a generator, we propose to produce a class-agnostic foreground map using BAS and further combine it with the class-specific maps to improve the quality of the initial seed, see Fig.~\ref{fig:foreground} and Table~\ref{table:foreground}. \textbf{5)} To further improve the segmentation quality, we propose to apply the losses of BAS as evaluation scores in the inference phase to assess each threshold and find the image-specific threshold on WSSS, see Fig.~\ref{fig:evaluate} and Table~\ref{table:evaluate}. \textbf{6)} We have made a lot of efforts to improve the presentations (\eg, motivation, related illustrative diagrams, formulation, experimental analysis, key results), and organizations of our paper. Besides, several sections have been refined to improve the readability and provide more detailed explanations about the motivation, quantitative/qualitative comparisons, and discussions.

The rest of this paper is organized as follows. Section~\ref{sec:related work} describes existing works related to WSOL and WSSS. The detailed method is described in Section~\ref{sec:methodology}. Section~\ref{sec:WSOL_experiment} and Section~\ref{sec:WSSS_experiment} present the experimental results of WSOL and WSSS, respectively. Limitation and future work are discussed in Section~\ref{sec:discussion}. Finally, we conclude our work in Section~\ref{sec:conclusion}.

\section{Related Work}\label{sec:related work}
\subsection{Weakly Supervised Object Localization}
Weakly supervised object localization (WSOL) is a challenging task that requires localizing objects using only image-level labels. To obtain localization results from the classification network, CAM~\citep{zhou2016learning} proposes to replace top layers with a global average pooling, and multiply the fully connected weights on depth feature maps to generate class activation map (CAM) as the localization map. Unfortunately, CAM usually focuses on the most discriminative regions. To alleviate this problem, a series of methods propose to use erasing strategies. HaS~\citep{singh2017hide} splits the original image into different patches and randomly masks part of them, forcing the classification network to learn more features of objects. ACoL~\citep{zhang2018adversarial} and EIL~\citep{mai2020erasing} erase areas with high response in the feature map and use two parallel branches for adversarial erasing. Differently, ADL~\citep{choe2019attention} erases the most significant regions of each layer during forward propagation, to achieve a balance between classification and localization. CutMix~\citep{yun2019cutmix} adopts a data enhancement strategy that mixes two different images to force network to learn relevant regions of different objects.

In addition, another class of approaches adopt the thought of spreading confidence regions to mine relevant features. SPG~\citep{zhang2018self} uses thresholds to filter foreground and background regions with high confidence from CAM to guide shallow network learning. Further, SPOL~\citep{wei2021shallow} generates more reliable confidence regions by multiplicative feature fusion strategy and trains a full segmentation network with confidence regions as pseudo labels. I2C~\citep{zhang2020inter} proposes to increase the robustness and reliability of localization by considering the correlation of different pictures from the same class. Besides, SPA~\citep{pan2021unveiling} uses a post-processing approach to extract feature maps with structure-preserving. SLT~\citep{guo2021strengthen} considers several similar classes as one class when 
generating classification loss and localization maps, which alleviates the problem of focusing on the most discriminative regions by strengthening learning tolerance. DA-WSOL~\citep{zhu2022weakly} aligns the feature distributions between the image and pixel domains with the thought of domain adaptation.

Most recently, two Foreground-Prediction-Map-based works~\citep{xie2021online,meng2021foreground}, both achieve the localization task by generating a foreground prediction map. ORNet~\citep{xie2021online} uses a two-stage approach, where an encode-decode layer is inserted in the shallow layer of the network as a generator and trained by the classification task in the first stage. In the second stage, the parameters of the classification network are fixed as an evaluator, and the foreground prediction map output by the generator is used to mask the image. Then the masked image is fed into the evaluator for classification training, so that the foreground prediction map can learn the object region. FAM~\citep{meng2021foreground} utilizes a Foreground Memory Mechanism structure to store different foreground classifiers and generate a class-agnostic foreground prediction map. The foreground prediction map is split into several specific parts which are used to mask the feature map to obtain different part-aware feature maps. After classification training with the corresponding foreground classifiers, the class-agnostic foreground map is forced to learn different object regions. It can be noticed that both ORNet~\citep{xie2021online} and FAM~\citep{meng2021foreground} only consider foreground regions and use cross-entropy to facilitate the learning of generator. Different from these methods, this paper proposes a background activation suppression strategy to learn foreground prediction maps through a simple but effective approach.

\subsection{Weakly Supervised Semantic Segmentation }
Weakly supervised semantic segmentation (WSSS) purposes to alleviate the reliance on pixel-level ground-truth labels by using weak labels instead. Existing WSSS methods usually include the following three stages: 1) obtaining a high-quality initial seed. 2) seed refinement and generating pseudo labels. 3) training a full segmentation network with pseudo labels. It can be seen that generating a high-quality pixel-level localization map is also crucial for WSSS, similar to WSOL.

\myPara{Seed Generation. }Extraction of CAM is arguably the most common and convenient approach to generate the initial seed, despite the problem that only the discriminative regions can be highlighted. To alleviate this issue, some methods propose to improve the quality of CAM by iterative manipulation. AE-PSL~\citep{wei2017object} performs iterative training steps to mine more object-related regions with adversarial erasure. RIB~\citep{lee2021reducing} applies a post-processing method to fine-tune the classification model and obtain CAMs by iteration. AdvCAM~\citep{lee2022anti} proposes an anti-adversarial approach to continuously identify more object areas. Besides, a category of methods try to improve the classification learning process. CONTA~\citep{zhang2020causal} aims to avoid contextual confusion by proposing a structural causal model to analyze the causalities among images, contexts, and class labels. SEAM~\citep{wang2020self} applies consistency regularization on CAMs through various sized images to mitigate the supervision gap issue. ReCAM~\citep{chen2022class} proposes to use softmax cross-entropy loss to suppress the response of different categories to the same receptive field. CLIMS~\citep{xie2022clims} utilizes the CLIP~\citep{radford2021learning} model to assist the network in activating more complete object regions. GAIN~\citep{li2018tell} uses Grad-CAM to obtain localization maps and improve them by exploiting the prediction scores of the network as supervision. In contrast, BAS is based on the FPM-based paradigm and proposes a more essential and effective background activation suppression loss compared to the cross-entropy used in the FPM-based methods from the experimental observations.


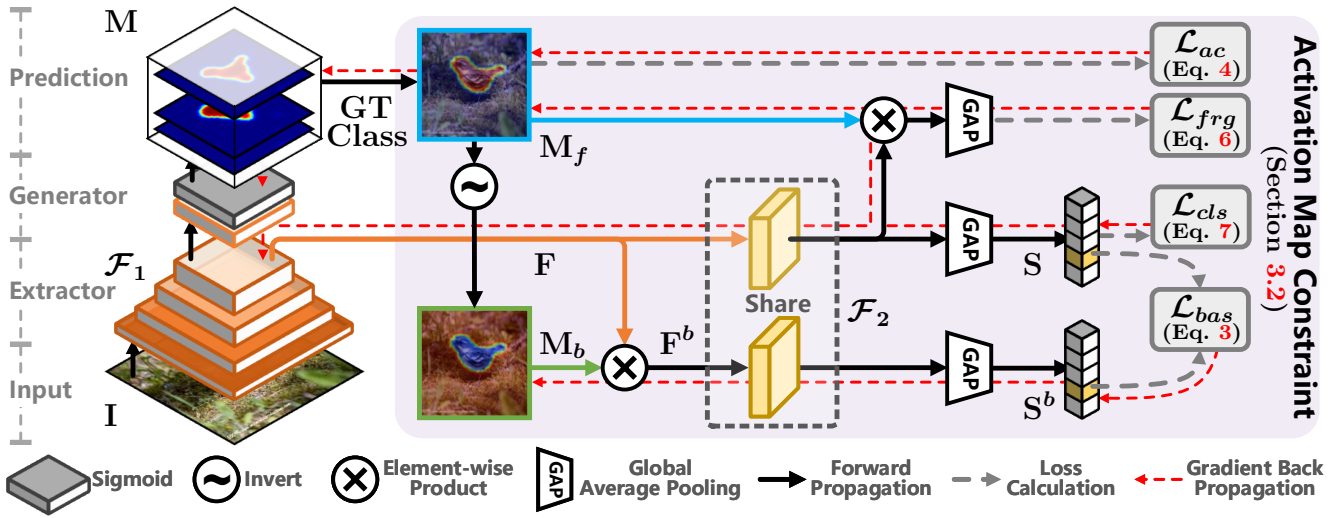
\begin{figure*}[t]
\centering
    \begin{overpic}[width=1.\linewidth]{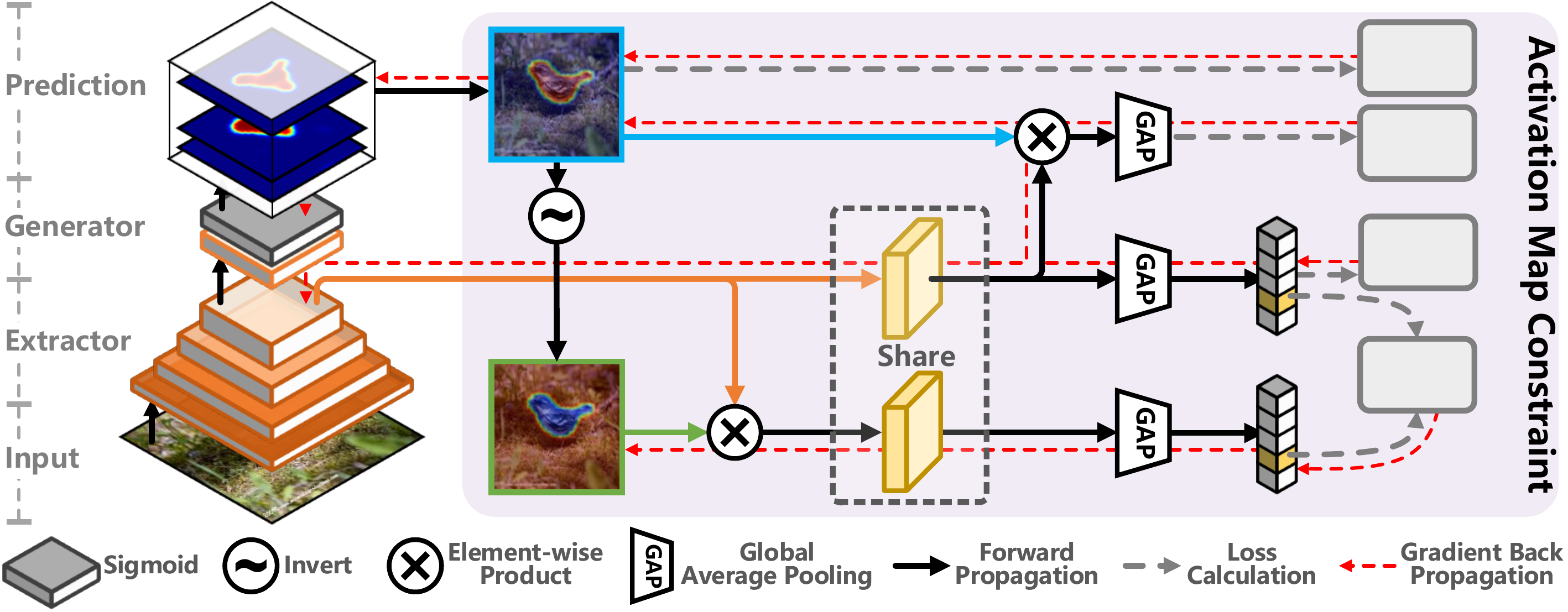}
        \put(7.5, 7.){\large\bm{{$\mathbf{I}$}}}
        \put(7.5,18.5){\large\bm{$\mathcal{F}_1$}}
        \put(40,18.5){\large\bm{$\mathbf{F}$}}
        \put(40.4,12.5){\large\bm{$\mathbf{M}_{b}$}}
        \put(40.4,27.5){\large\bm{$\mathbf{M}_{f}$}}
        \put(25.3, 30.5){\large\bm{{$\mathbf{GT}$}}}
        \put(24.3, 28.3){\large\bm{{$\mathbf{Class}$}}}
	    \put(7.5, 37){\large\bm{{$\mathbf{M}$}}}
	    \put(63.7, 15){\large\bm{$\mathcal{F}_2$}}
	    \put(49.5, 12.5){\large\bm{$\mathbf{F}^{b}$}}
	    \put(77, 7.5){\large\bm{$\mathbf{S}^{b}$}}
	    \put(77, 18.5){\large\bm{$\mathbf{S}$}}

	    \put(88.4, 35.5){\large\bm{$\mathcal{L}_{ac}$}}
	    \put(87.5, 33.7){\small\textbf{(Eq. \ref{eq:Area})}}
	    \put(88.1, 30.3){\large\bm{$\mathcal{L}_{frg}$}}
	    \put(87.5, 28.3){\small\textbf{(Eq. \ref{eq:FRG})}}
	    \put(88.2, 23.1){\large\bm{$\mathcal{L}_{cls}$}}
	    \put(87.5, 21.3){\small\textbf{(Eq. \ref{eq:CE})}}
	    \put(88.1, 15.3){\large\bm{$\mathcal{L}_{bas}$}}
	    \put(87.5, 13.6){\small\textbf{(Eq. \ref{eq:BAS})}}
	    
	    \put(95, 28.5){\normalsize\textbf{\rotatebox{270}{(Section \ref{AMC_section})}}}
    \end{overpic}
       \caption{\textbf{The architecture of the proposed Background Activation Suppression (BAS) in the training phase.} The class-specific foreground prediction map $\mathbf{M}_{f}$ and the coupled background prediction map $\mathbf{M}_{b}$ are obtained by the generator according to the ground-truth ($\mathbf{GT}$) class, and then fed into the Activation Map Constraint module together with the feature maps $\mathbf{F}$.}
    \label{fig:network}
\end{figure*}


\myPara{Mask Generation. }The initial seed is usually coarse and needs to be refined. Some researchers adopt the thought of region growing to spread the initial seed. SEC~\citep{kolesnikov2016seed} proposes three principles: seed, expand and constrain. The initial seed is expanded during the training of segmentation and constrained to the object boundaries. PSA~\citep{ahn2018learning} trains a deep network to predict semantic affinity between a pair of adjacent image coordinates and propagate the semantics by random walk~\citep{lovasz1993random}. IRN~\citep{ahn2019weakly} predicts a transition probability matrix from the boundary activation map and generates pseudo masks in a similar way to PSA.

\section{Methodology \label{sec:methodology}}
In this section, we first introduce the main architecture of the network and the definition of the symbols in Section~\ref{overview}. Then we describe the structure of the AMC module, including the form of the four loss functions, and the improvement of BAS compared to the previous conference version in Section~\ref{AMC_section}. The total loss functions for WSOL and WSSS are listed in Section~\ref{sec:WSOL} and Section~\ref{sec:WSSS}, respectively. Finally, we provide specific details of the exploratory experiments and statistical results on three different datasets in Section~\ref{sec:exploratory}. 

\subsection{Overview \label{overview}}
Based on the experimental observation, we enhance the completeness of the localization map for WSOL by proposing a background activation suppression (BAS) approach. As shown in Fig.~\ref{fig:network}, BAS consists of three modules: an extractor, a generator, and an Activation Map Constraint (AMC) module. The extractor is used to extract features related to classification and localization. The generator is to produce the foreground prediction maps. The AMC module is to promote the learning of extractor and generator through four kinds of losses.

Specifically, we divide the original backbone network into two sub-networks $\mathcal{F}_1$ and $\mathcal{F}_2$ according to the location of the generator, and denote the network parameter by $\Theta$. The sub-network $\mathcal{F}_1$ before the generator is used as a feature extractor. Given an image $\mathbf{I}$, the feature maps $\mathbf{F} \in \mathbb{R}^{H \times W \times N}$ are generated by extractor $\mathcal{F}_1(\mathbf{I}, \Theta_1)$ in the forward propagation, where $H$, $W$, and $N$ denote the height, width, and number of channels of the feature maps, respectively. Afterward, the feature maps $\mathbf{F}$ are fed into the generator, which consists of a $3\times3$ convolution layer and a \texttt{Sigmoid} activation function for generating a set of foreground prediction maps $\mathbf{M} \in \mathbb{R}^{H \times W \times C}$ with 0-1 distribution, where C is the number of categories. We choose the class-specific foreground prediction map $\mathbf{M}_{f} \in \mathbb{R}^{H \times W \times 1}$ corresponding to the ground-truth class and invert it to obtain the coupled background prediction map $\mathbf{M}_{b} \in \mathbb{R}^{H \times W \times 1}$, where $\mathbf{M}_{b}$ = $1 - \mathbf{M}_{f}$. Finally, $\mathbf{M}_{f}$, $\mathbf{M}_{b}$, and $\mathbf{F}$ are fed together into AMC module for prediction map learning. We will detail describe the AMC structure and loss functions in Section \ref{AMC_section}.

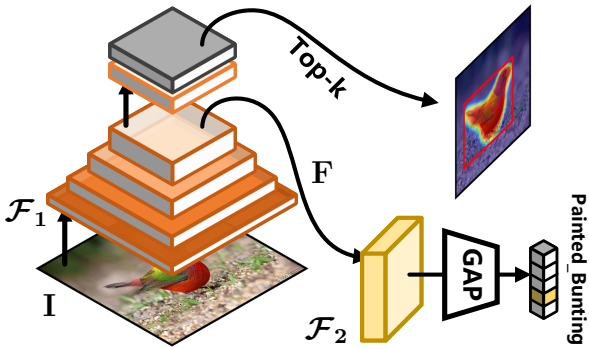
\begin{figure}[t]
\centering
    \begin{overpic}[width=1.\linewidth]{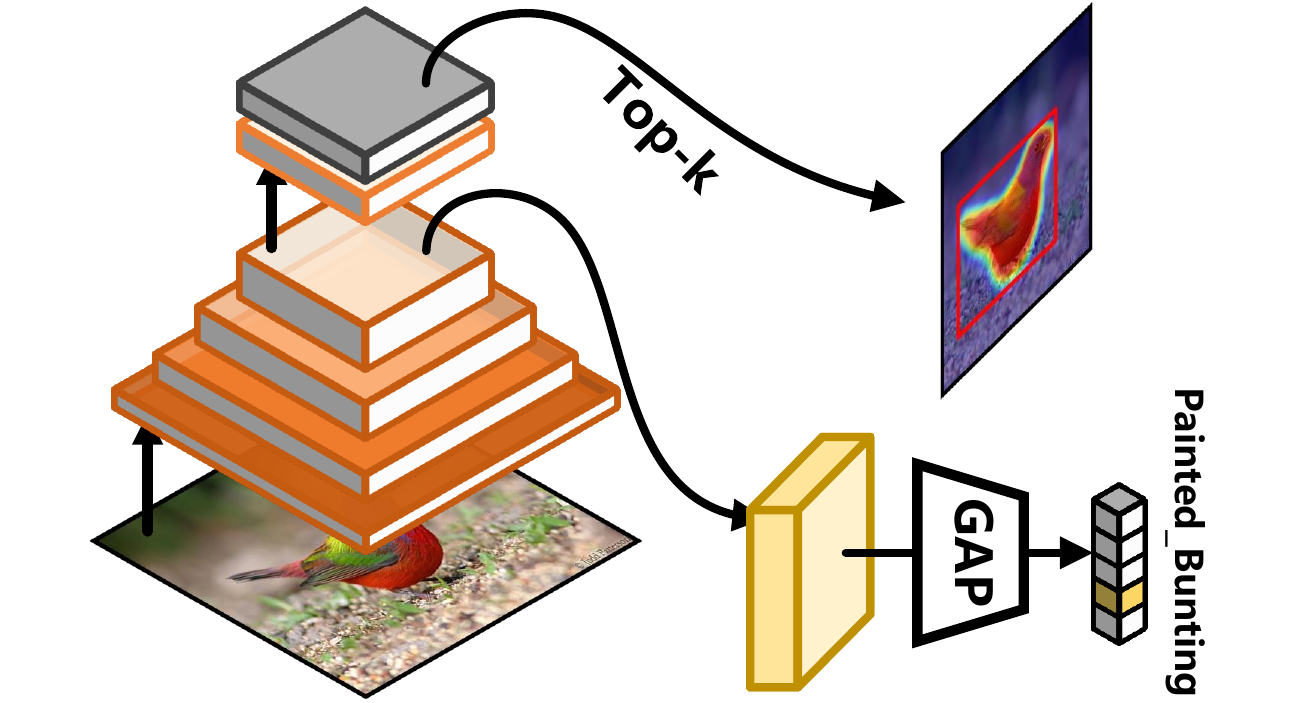}
        \put(8, 5){\large\bm{{$\mathbf{I}$}}}
	    \put(2, 20){\large\bm{$\mathcal{F}_1$}}
	    \put(49, 1.5){\large\bm{$\mathcal{F}_2$}}
	    \put(50, 26){\large\bm{$\mathbf{F}$}}
    \end{overpic}
       \caption{\textbf{The architecture of the proposed BAS in inference phase.} We utilize Top-$k$ to generate final localization map.}
    \label{fig:network_inf}
\end{figure}

In the inference phase, as illustrated in Fig.~\ref{fig:network_inf}, the feature maps $\mathbf{F}$ obtained by the extractor are input into the generator and sub-network $\mathcal{F}_2(\mathbf{F},\Theta_2)$ to generate the foreground prediction maps set $\mathbf{M}$ and the classification prediction logits $\mathbf{\tilde{y}}$, respectively. We select the prediction maps corresponding to the Top-$k$ predicted categories including the ground-truth class, and take their average values as the final localization result. Notably, the Top-$k$ strategy is only used in WSOL and not in WSSS.

\subsection{Activation Map Constraint \label{AMC_section}}
The proposed AMC module utilizes foreground map, background map, and feature maps as input to jointly promote the learning of extractor and generator, which is consisted of four different kinds of losses, including $\mathcal{L}_{bas}$, $\mathcal{L}_{ac}$, $\mathcal{L}_{frg}$, and $\mathcal{L}_{cls}$.

\myPara{Background Activation Suppression (\bm{$\mathcal{L}_{bas}$}).} For the input background prediction map $\mathbf{M}_{b}$, we multiply it by the feature maps $\mathbf{F}$ to obtain the background feature maps ($\mathbf{F} \cdot \mathbf{M}_{b}$), denoted as $\mathbf{F}^{b}\in \mathbb{R}^{H \times W \times N}$. Subsequently, the feature maps $\mathbf{F}$ and $\mathbf{F}^{b}$ are fed to two sub-networks $\mathcal{F}_2(\mathbf{F},\Theta_2)$ and $\mathcal{F}_2(\mathbf{F}^{b},\Theta_2)$ with shared weights, respectively. For the sub-network with $\mathbf{F}^{b}$ as input, the goal is to generate the background activation value by the same function, and the parameters of this sub-network are frozen in the back propagation. Following the sub-network $\mathcal{F}_2(\mathbf{F},\Theta_2)$ and the global average pooling (GAP)~\citep{zhou2016learning}, $\mathbf{F}$ and $\mathbf{F}^{b}$ produce the prediction logits $\mathbf{\tilde{y}}\in \mathbb{R}^{C}$ and $\mathbf{\tilde{y}}^{b}\in \mathbb{R}^{C}$, respectively, which can be expressed as follows:
\begin{align}
    &\mathbf{\tilde{y}}=\text{GAP}\left ( \mathcal{F}_2\left ( \mathbf{F},\Theta_2 \right ) \right ), \label{eq:pred_y} \\
    &\mathbf{\tilde{y}}^{b}=\text{GAP}\left ( \mathcal{F}_2\left ( \mathbf{F}^{b},\Theta_2 \right ) \right ). \label{eq:pred_y_b}
\end{align}
We select the values in the $\mathbf{\tilde{y}}$ and $\mathbf{\tilde{y}}^{b}$ according to the ground-truth class. After applying a \texttt{ReLU} activation function, these values are represented as the activation value $\mathbf{S}\in \mathbb{R}^{1}$ and the background activation value $\mathbf{S}^{b}\in \mathbb{R}^{1}$, respectively. $\mathbf{S}$ represents the activation value generated by the unmasked feature map, containing both foreground and background information, and $\mathbf{S}^{b}$ is the activation value generated by the background feature map, retaining only the background information. Here, we measure the difference between background activation value and activation value in a ratio form as a way to achieve background activation value suppression, and $\mathcal{L}_{bas}$ is defined as follows:
\begin{equation}
    \label{eq:BAS}
    \mathcal{L}_{bas}= \frac{ \mathbf{S}^{b} }{\mathbf{S}+ \varepsilon},
\end{equation}
where $\varepsilon$ is a very small value ($e^{-8}$), to ensure that the equation is meaningful. This ratio form not only avoids the addition of more hyperparameters, but also acts as a normalization, so that the range of loss value is maintained under an order of magnitude.

\begin{figure}[t]
\small
\centering
    \begin{overpic}[width=1.\linewidth]{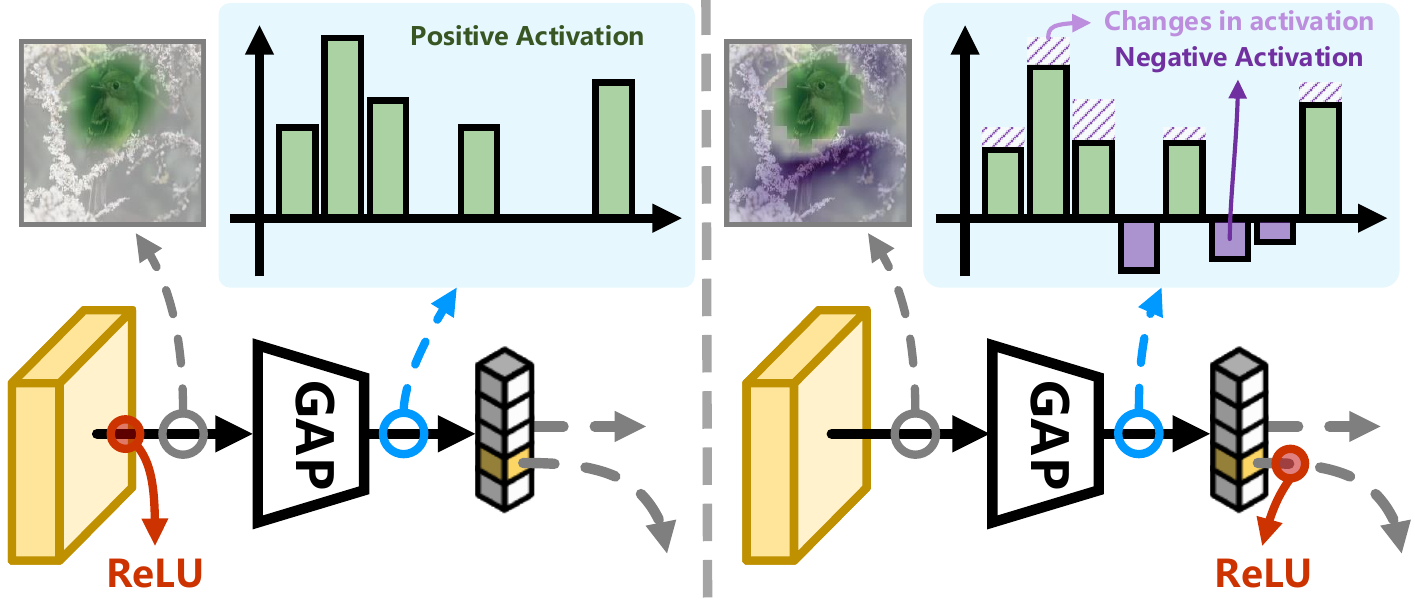}
    \put(23, -1){\textbf{(a)}}
    \put(72, -1){\textbf{(b)}}
    \put(93., 0.5){$\mathbf{S}$/$\mathbf{S}^{b}$}
    \put(38, 0.5){$\mathbf{S}$/$\mathbf{S}^{b}$}
    \end{overpic}
       \caption{\textbf{ The improvement of BAS.} Partial structure of \textbf{(a)} the previous conference version and \textbf{(b)} this work. The \textcolor{mygreen}{\textbf{green}} pixels in the localization map indicate positive values and the \textcolor{mypurple}{\textbf{purple}} ones indicate negative values.}
    \label{fig:relu}
\end{figure}
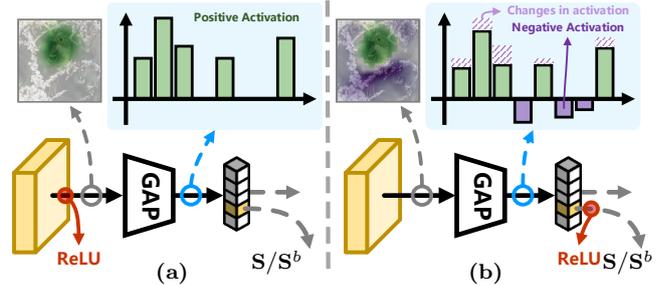

Generating a non-negative $\mathbf{S}$ and $\mathbf{S}^{b}$ is necessary for $\mathcal{L}_{bas}$. In the previous conference version, we use a \texttt{ReLU} as the activation function at the end of the network to ensure the non-negativity of the outputs, as shown in Fig.~\ref{fig:relu}. This approach causes pixels with negative values are marked as 0 after \texttt{ReLU} and their gradients will not take part in the back propagation. While pixels with negative values are usually associated with background areas, which are also important for the learning of classification and prediction maps. As shown in Fig. \ref{fig:bas_loss}, the neglect of negative activation values in the classification loss indirectly causes the BAS loss to become inadequate (the loss value becomes larger instead) later in the training process. To solve this problem, we remove this \texttt{ReLU} layer to make negative pixels also participate in the gradient back propagation. To ensure the non-negativity of $\mathbf{S}$ and $\mathbf{S}^{b}$, we use the \texttt{ReLU} activation function separately before generating them.

\myPara{Area Constraint (\bm{$\mathcal{L}_{ac}$}).} The background prediction map can be guided by $\mathcal{L}_{bas}$ in a suppressed way, and a smaller $\mathcal{L}_{bas}$ means that the region covered by the background prediction map is less discriminative. When the background prediction map can cover the background region well, the $\mathcal{L}_{bas}$ it produced has to be minimal while the background area should be as large as possible, accordingly, the foreground area should be as small as possible. So we use the foreground prediction map area as constraints:
\begin{equation}
    \label{eq:Area}
    \mathcal{L}_{ac}= \frac{1}{H \times W}\sum_{h=1}^H \sum_{w=1}^W \mathbf{M}_{f}\left ( h,w \right ).
\end{equation}

\myPara{Foreground Region Guidance (\bm{$\mathcal{L}_{frg}$}).} Meanwhile, we maintain the FPM's approach of employing the classification task to drive the learning of foreground prediction maps, which uses high-level semantic information to guide the foreground prediction map to the approximate correct region of the object. Consequently, a foreground region guidance loss based on cross-entropy is utilized. After $\mathbf{F}$ is fed into $\mathcal{F}_2(\mathbf{F},{\Theta}_2)$, it is dotted with $\mathbf{M}_{f}$ to produce $\mathcal{L}_{frg}$:
\begin{align}
   &\mathbf{\tilde{y}}^{f}=\text{GAP}\left ( \mathbf{M}_{f} \cdot \mathcal{F}_2\left ( \mathbf{F},\Theta_2 \right ) \right ), \label{eq:pred_y_f} \\
   &\mathcal{L}_{frg}=-\sum_{i=1}^C \mathbf{y}_{i} \log_{}{\frac{e^{\mathbf{\tilde{y}}_{i}^{f}}}{\sum_{j}^{C} e^{\mathbf{\tilde{y}}_{j}^{f}}}}, \label{eq:FRG}
\end{align}
where $\mathbf{y}$ denotes the image-level one-hot encoding label. 

\myPara{Classification (\bm{$\mathcal{L}_{cls}$}).} Besides, we obtain the classification loss $\mathcal{L}_{cls}$ by applying cross-entropy to $\mathbf{\tilde{y}}$, which is used for classification learning of the entire image:
\begin{equation}
    \label{eq:CE}
   \mathcal{L}_{cls}=-\sum_{i=1}^C \mathbf{y}_{i} \log_{}{\frac{e^{\mathbf{\tilde{y}}_{i}}}{\sum_{j}^{C} e^{\mathbf{\tilde{y}}_{j}}}}.
\end{equation}
\subsection{Weakly Supervised Object Localization \label{sec:WSOL}}
By jointly optimizing background activation suppression loss, area constraint loss, foreground region guidance loss, and classification loss in the AMC module, the foreground prediction map can be guided to the overall area of the object. The total loss of the BAS training process is defined in the following form:
\begin{equation}
    \label{eq:WSOL}
    \mathcal{L}=\mathcal{L}_{cls} + \alpha\mathcal{L}_{frg} + \beta\mathcal{L}_{ac} + \lambda\mathcal{L}_{bas},
\end{equation}
where $\alpha$, $\beta$, and $\lambda$ are hyperparameters, $\mathcal{L}_{cls}$ and $\mathcal{L}_{frg}$ are both cross-entropy losses. For all backbones and datasets, we set $\lambda =1$. The ablation experiments of the hyperparameters $\alpha$, $\beta$, and $\lambda$ on WSOL are described in Section \ref{ablation_section}.

\subsection{Weakly Supervised Semantic Segmentation \label{sec:WSSS}}

BAS can also be applied to weakly supervised semantic segmentation to verify the generality of our method. Different from weakly supervised object localization, weakly supervised semantic segmentation no longer assumes that there is only one ground-truth class in an image, which is more challenging. In addition, it is more direct to reflect the segmentation quality of the prediction map by comparing with the weakly supervised semantic segmentation SOTA methods.

\begin{figure}[t]
\centering
    \begin{overpic}[width=1.\linewidth]{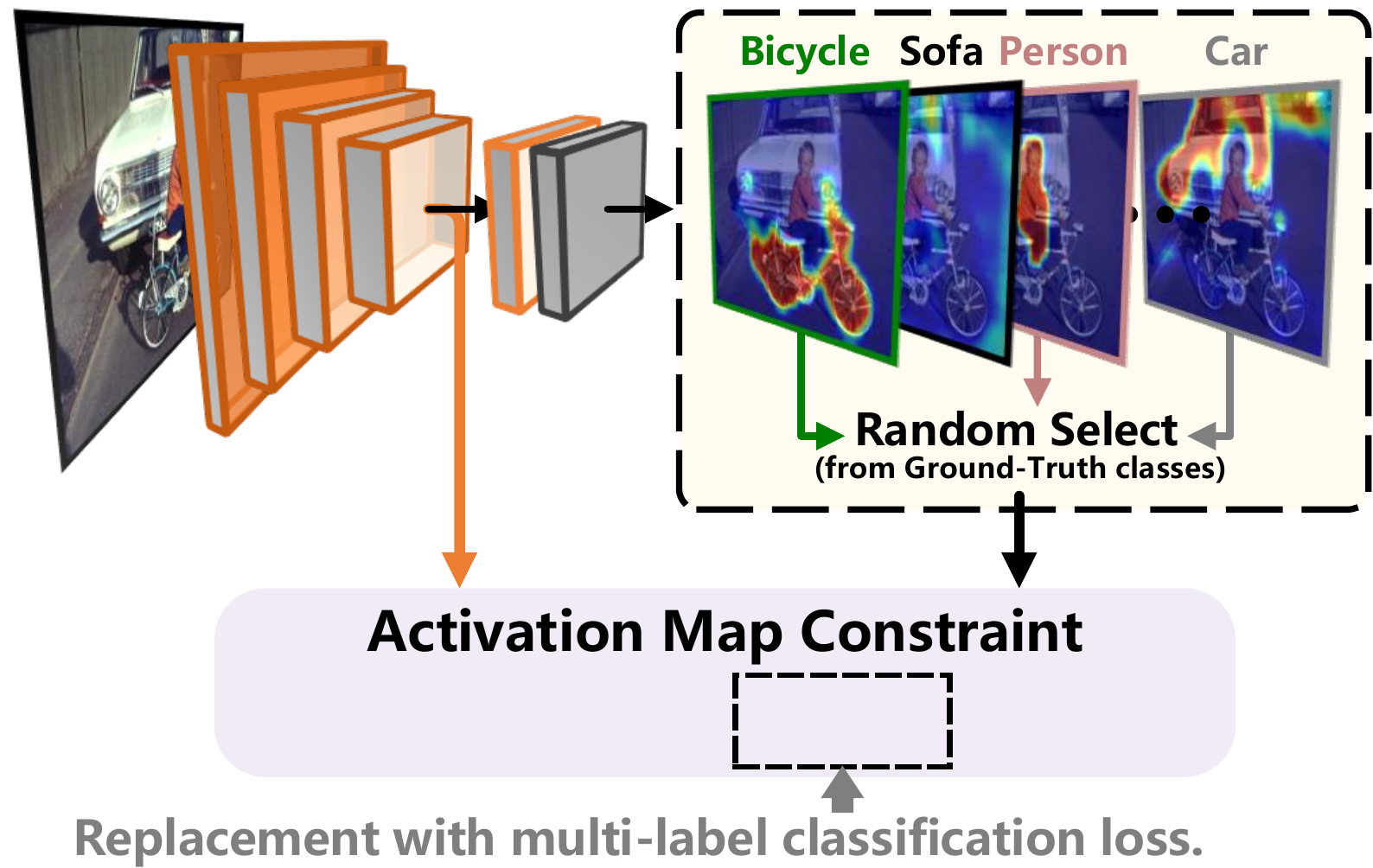}
    \put(23, 9.){\Large\bm{$\left ( \mathcal{L}_{ac}, \mathcal{L}_{frg}, \mathcal{L}_{mcls}, \mathcal{L}_{bas} \right )$}}
    \put(89, 1.5){\small\textbf{(Eq. \ref{eq:mcls})}}
    \end{overpic}
       \caption{\textbf{Applying BAS to weakly supervised semantic segmentation task.} }
    \label{fig:network_wsss}
\end{figure}

Based on the network structure in Fig.~\ref{fig:network}, we apply BAS to weakly supervised semantic segmentation with minor changes. As shown in Fig.~\ref{fig:network_wsss}, we maintain the learning process for a single prediction map in the AMC module by randomly selecting a foreground category in the image and denoting its corresponding prediction map as $\mathbf{M}_{f}$. In addition, to make the network achieve multi-label classification, we adopt softmax cross-entropy loss and simply modify the form of it instead of using \texttt{Sigmoid}-based loss (binary cross-entropy loss). It mainly due to the activation value $\mathbf{S}^{b}$ obtained from the background localization map has to be less than 0 to ensure that the probability generated by $1/({1+e^{-\mathbf{S}^{b}}})$ is close to 0, which conflicts with the non-negativity of $\mathbf{S}^{b}$.

\myPara{Multi-Label-Classification (\bm{$\mathcal{L}_{mcls}$}).} For weakly supervised semantic segmentation task, we adopt the multi-label classification loss $\mathcal{L}_{mcls}$ instead of $\mathcal{L}_{cls}$ to deal with the multi-label case. To avoid the problems of class imbalance and training instability when there are multi-label in the softmax formulation, we only consider the differentiation between foreground and background classes and ignore the interrelationship among foreground categories. It can be expressed as follows:
\begin{equation}
   \mathcal{L}_{mcls}=-\sum_{i=1}^{L} \mathbf{y}_{i} \log_{}{\left ( \frac{e^{\mathbf{\tilde{y}}_{i}}}{{\sum_{j}^{K} e^{\mathbf{\tilde{y}}_{j}}} + e^\mathbf{\tilde{y}_{i}}} \right )}, \label{eq:mcls}
\end{equation}
where $L$ is the set of ground-truth classes in the image, and the remaining set of categories is denoted as $K$.
The total loss function in weakly supervised semantic segmentation is of the following form:
\begin{equation}
    \mathcal{L}=\mathcal{L}_{mcls} + \alpha\mathcal{L}_{frg} + \beta\mathcal{L}_{ac} + \lambda\mathcal{L}_{bas}.
\end{equation}
The $\lambda$ is set to 1 for all datasets. For \texttt{PASCAL VOC 2012}, we set $\alpha=0.2$ and $\beta=1.2$. For \texttt{MS COCO 2014}, we adopt $\alpha=0.5$ and $\beta=1.5$. The ablation experiments of the hyperparameters $\alpha$, $\beta$, and $\lambda$ on WSSS, and the results of different combinations of hyperparameters on five datasets are presented in Section~\ref{ablation_section_WSSS}. 

\begin{figure}[t]
\small
\centering
    \begin{overpic}[width=1.\linewidth]{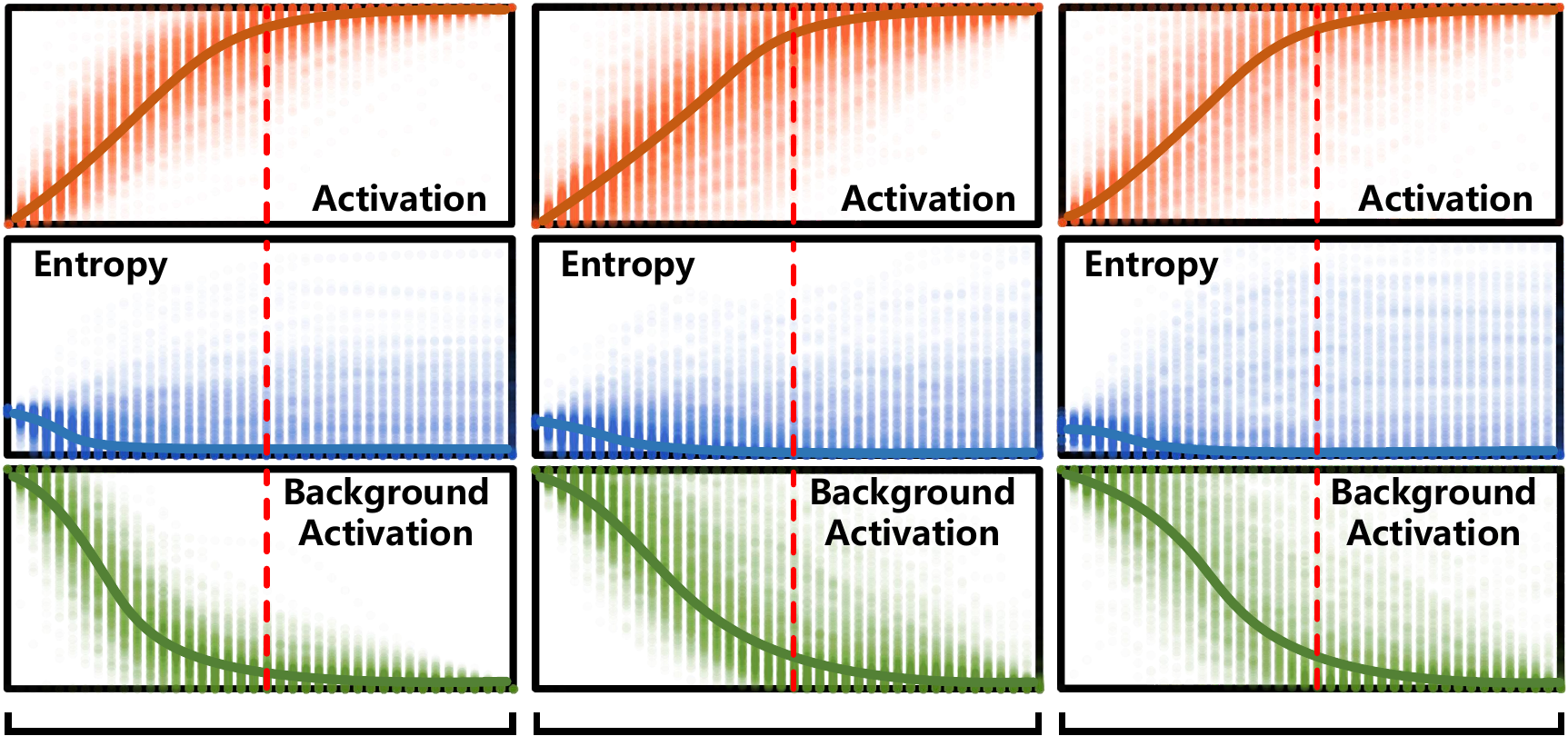}
        \put(5., -0.3){\colorbox{white}{\textbf{\texttt{CUB-200-2011}}}}
    	\put(40.5, -0.3){\colorbox{white}{\textbf{\texttt{OpenImages}}}}
    	\put(69., -0.3){\colorbox{white}{\textbf{\texttt{PASCAL VOC 2012}}}}
    \end{overpic}
       \caption{\textbf{Motivation. Statistical analysis about exploratory experiments on different datasets.}}
    \label{fig:basexp}
\end{figure}

\subsection{Empirical Justification \label{sec:exploratory}}

In this part, we empirically justify the advantage of introducing background activation suppression and its generalizability.

The purpose of the exploratory experiment is to investigate the relationship between activation value (\textbf{Activation}), cross-entropy (\textbf{Entropy}) and background activation value (\textbf{Background Activation}) with the mask area. Specifically, we first train a VGG16 classification network on \texttt{CUB-200-2011} using $\mathcal{L}_{cls}$ (Equation~\ref{eq:CE}) as supervision. Then, for a given pixel-level mask, the activation and entropy corresponding to this mask are generated by masking the feature map. We erode and dilate the ground-truth mask with a convolution of kernel size $5n \times 5n$, obtain masks with different areas by changing the value of $n$, and plot the activation versus entropy with the mask area as the horizontal axis. As shown in Fig.~\ref{fig1} (A), we display the curve for a single image through the above process.

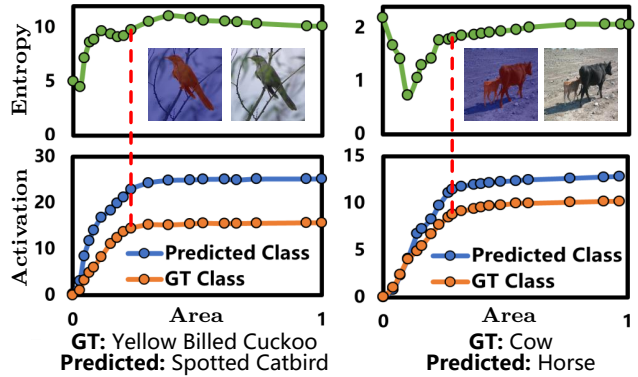
\begin{figure}[t]
\small
\centering
    \begin{overpic}[width=1.\linewidth]{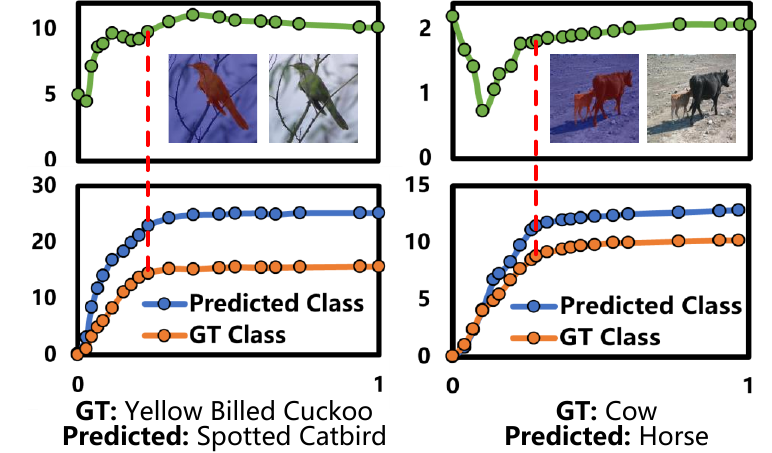}

    \put(0.5, 41){\rotatebox{90}{\textbf{Entropy}}}
    \put(0.5, 14){\rotatebox{90}{\textbf{Activation}}}
    
    \put(25., 8.5){\textbf{Area}}
    \put(74, 8.5){\textbf{Area}}
    \end{overpic}
    
    \caption{\textbf{Cross-entropy presents a divergence trend} as the area of the mask increases when the model classifies the object incorrectly. The dashed line represents the position of ground-truth mask. Entropy: cross-entropy. GT: Ground-Truth.}
    \label{fig:diverged}
\end{figure}

Due to each image having a different activation value distribution and a different ground-truth mask area, we normalize the activation curve for each image by dividing the activation value generated by the entire image to obtain a more statistically significant result, the same as in Equation~\ref{eq:BAS}. In addition, the area representing the horizontal axis is also normalized based on the ground-truth mask area, which is marked by a red line. As shown in Fig.~\ref{fig:basexp}, we present the curves of foreground activation value, cross-entropy, and background activation value with respect to the mask area, which are counted on the \texttt{CUB-200-2011} test set. It can be noted that the samples on the whole present the following phenomena: 
When the mask expands near the ground-truth mask, the activation value starts to saturate and the corresponding background activation value tends to converge, while cross-entropy converges to zero early or even diverges with the expansion of the mask. This suggests that the object region learned by activation values is larger and closer to the real object region than that learned by cross-entropy. We further explore why the cross-entropy occasionally diverges and visualize some results as shown in Fig.~\ref{fig:diverged}. It can be noted that when the network classifies objects incorrectly, such as identifying cows as horses, the calculated cross-entropy maintains a high value as the mask area increases. In this case, adopting cross-entropy values to supervise the localization map is less feasible and appropriate than using activation values which are not influenced by other categories. Besides, to verify the generality of this observation, we perform the same experiments on the more complex \texttt{OpenImages} and \texttt{PASCAL VOC 2012} datasets. For \texttt{PASCAL VOC 2012}, we select one ground-truth category and its corresponding mask at a time, convert the multi-label into single-label, and then plot the curve in the same way. As shown in Fig.~\ref{fig:basexp}, the statistical analysis demonstrates similar phenomena, therefore, we believe it is general that better localization ability can be learned through activation values compared to cross-entropy.

\begin{table*}[t]
\caption{\textbf{Comparison with state-of-the-art methods.} Best results are highlighted in \textbf{bold}, second are \underline{underlined}. ${\lozenge}$ means multi-stage model. $\times n$ means that there are $n$ different networks used.}
\renewcommand{\arraystretch}{1.}
\renewcommand{\tabcolsep}{2pt}
\small
\centering
\begin{tabular}{l|c|c|ccc|ccc}
\Xhline{2.0\arrayrulewidth}
\hline
\multirow{2}{*}{\textbf{Methods}} & \multirow{2}{*}{\textbf{Venue}} & \multirow{2}{*}{\textbf{Backbone}} & \multicolumn{3}{c|}{\textbf{\texttt{CUB-200-2011} Loc Acc.}} & \multicolumn{3}{c}{\textbf{\texttt{ILSVRC} Loc Acc.}} \\
\cline{4-9}
      & &  & \textbf{Top-1 Loc} & \textbf{Top-5 Loc} & \textbf{GT-k.}  & \textbf{Top-1 Loc} & \textbf{Top-5 Loc} & \textbf{GT-k.} \\
\hline
\Xhline{2.0\arrayrulewidth}
CAM~\citep{zhou2016learning} & CVPR$16$ &VGG16
& 41.06 & 50.66 & 55.10 & 42.80 & 54.86 & 59.00  \\
ACoL~\citep{zhang2018adversarial} & CVPR$18$&VGG16
& 45.92 & 56.51 & 62.96 & 45.83 & 59.43 & 62.96 \\
ADL~\citep{choe2020attention} & TPAMI$20$ & VGG16
& 52.36 & $-$ & 75.41 & 44.92 & $-$ & $-$ \\
I2C~\citep{zhang2020inter} & ECCV$20$ &VGG16
& 55.99 & 68.34 & $-$ & 47.41 & 58.51 & 63.90 \\
MEIL~\citep{mai2020erasing} & CVPR20 &VGG16
& 57.46 & $-$ & 73.84 & 46.81 & $-$ & $-$ \\
PSOL~\citep{zhang2020rethinking} ${\lozenge}$ & CVPR$20$ &VGG16 $\times$2
& 66.30 & 84.05 & 89.11 & 50.89 & 60.90 & 64.03 \\
 SPA~\citep{pan2021unveiling} & CVPR$21$ &VGG16
& 60.27 & 72.50 & 77.29 & 49.56 & 61.32 & 65.05 \\
SLT~\citep{guo2021strengthen} ${\lozenge}$ & CVPR$21$ &VGG16 $\times$3  
& 67.80 & $-$ & 87.60 & 51.20 & 62.40 & 67.20 \\
FAM~\citep{meng2021foreground} & ICCV$21$ &VGG16
& 69.26 & $-$ & 89.26 & 51.96 & $-$ &\textbf{71.73}\\
ORNet~\citep{xie2021online} ${\lozenge}$ & ICCV$21$ &VGG16 $\times$2  
& 67.73 & 80.77 & 86.20 & 52.05 & 63.94 & 68.27 \\
Kim et al.~\citep{kim2022bridging} & CVPR$22$ &VGG16 
&\underline{70.83}&\textbf{88.07}& \textbf{93.17} & 49.94 & 63.25 & 68.92 \\
CREAM~\citep{xu2022cream} ${\lozenge}$ & CVPR$22$ & VGG16 $\times$2  
& 70.44 &\underline{85.67} & 90.98 & \underline{52.37} & \underline{64.20} & 68.32 \\
\hline
\rowcolor{mygray}
\textbf{BAS (ours)}&This Work &VGG16&
\textbf{70.90} & 85.36 &\underline{91.04} & \textbf{52.94} & \textbf{65.38} & \underline{69.66}\\
\hline
\Xhline{1.5\arrayrulewidth}
CAM~\citep{zhou2016learning} & CVPR$16$  &MobileNetV1
& 48.07 &\underline{$59.20$}& 63.30 & 43.35 & \underline{54.44} & 58.97 \\
HaS~\citep{singh2017hide} & ICCV$17$ &MobileNetV1
& 46.70 & $-$& 67.31 & 42.73 &$-$& 60.12 \\
ADL~\citep{choe2020attention} & TPAMI$20$ &MobileNetV1
& 47.74 & $-$ & $-$ & 43.01 &$-$&$-$\\
RCAM~\citep{bae2020rethinking} & ECCV$20$ &MobileNetV1
& 59.41 & $-$ & 78.60 & 44.78 & $-$ & 61.69 \\
FAM~\citep{meng2021foreground} & ICCV$21$ &MobileNetV1
&\underline{65.67} & $-$ & \underline{85.71} & \underline{46.24}& $-$ & \underline{62.05}\\
\hline
\rowcolor{mygray}
\textbf{BAS (ours)} & This Work & MobileNetV1
& \textbf{70.54}& \textbf{86.71}&\textbf{93.04} & \textbf{53.05}& \textbf{66.68}&\textbf{72.03}\\
\hline
\Xhline{1.5\arrayrulewidth}
CAM~\citep{zhou2016learning} & CVPR$16$ &ResNet50
& 46.71 & 54.44 & 57.35 & 48.69 & 58.00 & 60.58 \\
ADL~\citep{choe2020attention} & TPAMI$20$ &ResNet50
& 62.29 & $-$ & $-$ & 48.53 & $-$ & $-$ \\
PSOL~\citep{zhang2020rethinking} ${\lozenge}$ & CVPR$20$ &ResNet50 $\times$2
& 70.68 & 86.64 & 90.00 & 53.98 & 63.08 & 65.44 \\
FAM~\citep{meng2021foreground} & ICCV$21$ &ResNet50
& 73.74 &$-$& 85.73 & 54.46 &$-$& 64.56 \\
DA-WSOL~\citep{zhu2022weakly} & CVPR$22$ &ResNet50 $\times$2
& 66.65 & $-$ & 81.83 & \underline{55.84} & $-$ & \underline{70.27} \\
Kim et al.~\citep{kim2022bridging} & CVPR$22$ &ResNet50
& 73.16 & \underline{86.68} & \underline{91.60} & 53.76 & \underline{65.75} & 69.89 \\
CREAM~\citep{xu2022cream} ${\lozenge}$ & CVPR$22$ &ResNet50 $\times$2
& \underline{76.03} & $-$ & 89.88 & 55.66 & $-$ &  69.31 \\
\hline
\rowcolor{mygray}
$\textbf{BAS (ours)}$ & This Work &ResNet50
&\textbf{76.75}&\textbf{90.04}&\textbf{95.41} &\textbf{57.46}&\textbf{68.57}&\textbf{72.00}\\
\hline
\Xhline{1.5\arrayrulewidth}
CAM~\citep{zhou2016learning} & CVPR$16$ &InceptionV3
& 41.06 & 50.66 & 55.10  & 46.29 & 58.19 & 62.68 \\
DANet~\citep{xue2019danet} & ICCV$19$ &InceptionV3
& 49.45 & 60.46 & 67.03 & 47.53 & 58.28 & $-$ \\
I2C~\citep{zhang2020inter} & ECCV$20$ &InceptionV3
& 55.99 & 68.34 & 72.60 & 53.11 & 64.13 & 68.50 \\
GCNet~\citep{lu2020geometry} & ECCV$20$ &InceptionV3
& 58.58 & 71.00 & 75.30  & 49.06 & 58.09 & $-$\\
PSOL~\citep{zhang2020rethinking} ${\lozenge}$ & CVPR$20$ &InceptionV3 $\times$2
& 65.51 & 83.44 & $-$ & 54.82 & 63.25 & 65.21 \\
SPA~\citep{pan2021unveiling} & CVPR$21$ &InceptionV3
& 53.59 & 66.50 & 72.14 & 52.73 & 64.27 & 68.33 \\
SLT~\citep{guo2021strengthen} ${\lozenge}$ & CVPR$21$&InceptionV3 $\times$3
& 66.10 & $-$ & 86.50 & 55.70 & 65.40 & 67.60 \\
FAM~\citep{meng2021foreground} & ICCV$21$ &InceptionV3
& 70.67 & $-$ & 87.25 & 55.24 & $-$ & 68.62 \\
CREAM~\citep{xu2022cream} ${\lozenge}$ & CVPR$22$ &InceptionV3 $\times$2 
&\underline{71.76}&\underline{86.37}&\underline{90.43} &\underline{56.07}&\underline{66.19}&\underline{69.03}\\
\hline
\rowcolor{mygray}
$\textbf{BAS (ours)}$ & This Work &InceptionV3
&\textbf{72.09}&\textbf{88.11}&\textbf{94.63} &\textbf{58.50}&\textbf{69.03}&\textbf{72.07}\\
\hline
\Xhline{2.0\arrayrulewidth}
\end{tabular}
\label{table:comparsion}
\end{table*}

\begin{figure*}[t]
\centering
	\begin{overpic}[width=0.98\linewidth]{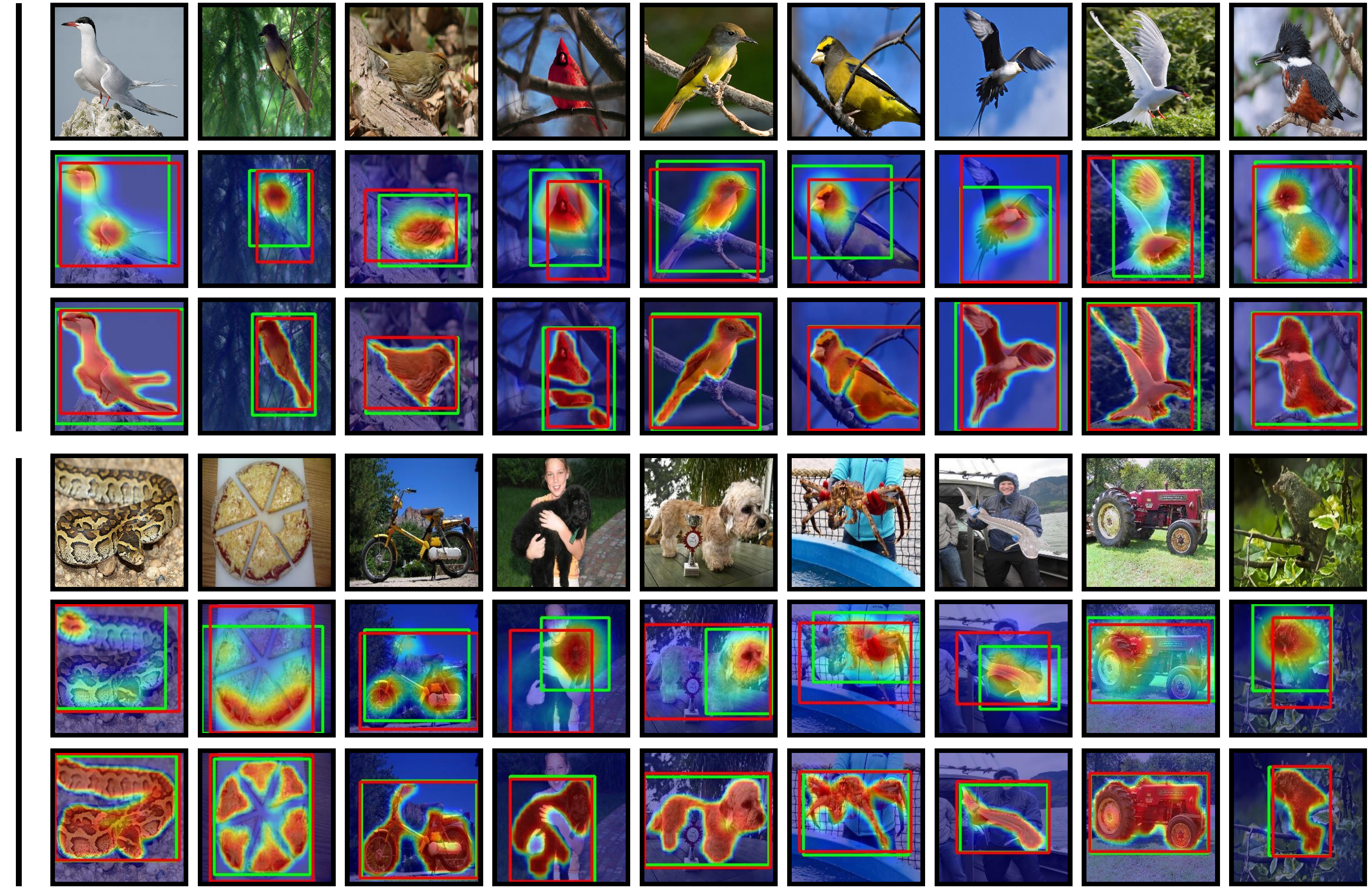}
	    \put(2, 3.7){\rotatebox{90}{\small\textbf{Ours}}}
	    \put(2, 13.5){\rotatebox{90}{\small\textbf{CAM}}}
	    \put(2, 24.2){\rotatebox{90}{\small\textbf{Image}}}
	    
	    \put(2, 36.5){\rotatebox{90}{\small\textbf{Ours}}}
	    \put(2, 46.5){\rotatebox{90}{\small\textbf{CAM}}}
	    \put(2, 57.){\rotatebox{90}{\small\textbf{Image}}}
	    
	    \put(-0.7, 43.5){\rotatebox{90}{\small\textbf{\texttt{CUB-200-2011}}}}
	    \put(-0.7, 13.5){\rotatebox{90}{\small\textbf{\texttt{ILSVRC}}}}
\end{overpic}
\caption{\textbf{Visualization comparison with the baseline CAM~\citep{zhou2016learning} method on \texttt{CUB-200-2011} and \texttt{ILSVRC}. The ground-truth bounding boxes are in \textcolor{red}{\textbf{Red}}, and the predictions are in \textcolor{green}{\textbf{Green}}.}}
\label{visual}
\end{figure*}

\section{Experiments on Weakly Supervised Object Localization \label{sec:WSOL_experiment}}
\subsection{Experimental Setup}

\myPara{Datasets.} We evaluate the proposed method on the popular benchmarks including \textbf{\texttt{CUB-200-2011}}~\citep{welinder2011caltech}, \textbf{\texttt{ILSVRC}}~\citep{russakovsky2015imagenet}, and \textbf{\texttt{OpenImages}}~\citep{choe2020evaluating}. \texttt{CUB-200-2011} contains 200 fine-grain classes of birds with 5,994 training images and 5,794 testing images. \texttt{ILSVRC} contains about 1.2 million training images and 50,000 validation images, which are divided into 1,000 categories. \texttt{OpenImages} consists of 29,819, 2,500 and 5,000 samples from 100 classes for training, validation and test, respectively. Except for class labels, \texttt{CUB-200-2011} and \texttt{OpenImages} also provide pixel-level mask annotations for the evaluation of the prediction mask.


\myPara{Metrics.} Following DA-WSOL~\citep{zhu2022weakly}, we apply both bounding box and mask metrics to evaluate the performance of our BAS. For bounding box, following~\cite{wu2023spatial,selvaraju2020grad,lee2022anti}, four metrics are used for evaluation, including GT-known localization accuracy (\textbf{GT-known Loc}), Top-1 localization accuracy (\textbf{Top-1 Loc}), Top-5 localization accuracy (\textbf{Top-5 Loc}), and maximal box accuracy (\textbf{MaxBoxAccV2}). Specifically, GT-known Loc is correct when the intersection over union (\textbf{IoU}) between the ground-truth bounding box and the predicted bounding box is greater than a fixed IoU threshold ($\delta$ = 0.5). Top-1/Top-5 Loc is correct when the Top-1/Top-5 predicted categories contain the ground-truth class and the GT-known Loc is correct. MaxBoxAccV2 compared to GT-known ($\delta$ = 0.5) considers multiple IoU thresholds ($\delta$ $\in$ \{0.3, 0.5, 0.7\}) and takes the average localization performance as the result. For mask, we adopt both the peak intersection over union (\textbf{PIoU})~\citep{zhang2020rethinking} and the pixel average precision (\textbf{PxAP})~\citep{choe2020evaluating} as metrics when the pixel-level ground-truth label is available.

\myPara{Implementation Details.} We evaluate the proposed method on the most popular backbones, including VGG16~\citep{simonyan2014very}, InceptionV3~\citep{szegedy2016rethinking}, ResNet50~\citep{he2016deep}, and MobileNetV1~\citep{howard2017mobilenets}. All networks are fine-tuned on the pre-trained weights of \texttt{ILSVRC}~\citep{russakovsky2015imagenet}. We train 120 epochs on the \texttt{CUB-200-2011}~\citep{welinder2011caltech} and 9 epochs on \texttt{ILSVRC}~\citep{russakovsky2015imagenet}. In the training phase, the input images are resized to 256$\times$256 and then randomly cropped to 224$\times$224. When $\mathcal{L}_{bas}$ is larger than 1, we mark it as 1, to ensure the stability of the initial training. In the inference phase, we use ten crop augmentation to get the final classification results following the settings in~\cite{pan2021unveiling,guo2021strengthen,zhang2018self}. For localization, we replace the random crop with the center crop, as in previous works~\citep{wei2021shallow,zhang2020rethinking,yun2019cutmix,choe2019attention}.

\begin{table*}[t]
    \centering
    \begin{minipage}{0.22\textwidth}
    \centering
    \caption{\textbf{Ablation study.} \textbf{(a)} the baseline method. \textbf{(b)} add $\mathcal{L}_{bas}$ to the baseline. \textbf{(c)} synthesize the prediction maps with Top-$k$ strategy.}
    \label{ablation}
    \end{minipage}
    \centering
    \begin{minipage}{0.40\textwidth}
    \renewcommand{\arraystretch}{1.}
    \renewcommand{\tabcolsep}{10pt}
    \small
    \centering
    \begin{tabular}{c|c|c|c}
    \hline
    \Xhline{2.\arrayrulewidth}
             & \textbf{(a)}   & \textbf{(b)}   & \textbf{(c)}   \\
    \hline
    \Xhline{2.\arrayrulewidth}
    \textbf{Baseline} & \checkmark     & \checkmark     & \checkmark     \\
    \bm{$\mathcal{L}_{bas}$}        &       & \checkmark     & \checkmark     \\
    \textbf{Top-\bm{$k$}}    &       &       & \checkmark     \\
    \hline
    \Xhline{2.\arrayrulewidth}
    \textbf{Top-1 Loc}    & 57.89 & 74.15 & \cellcolor{mygray}\textbf{76.75} \\
    \textbf{Top-5 Loc}    & 67.48 & 86.88 & \cellcolor{mygray}\textbf{90.04} \\
    \textbf{GT-known}     & 71.14 & 92.15 & \cellcolor{mygray}\textbf{95.41} \\
    \hline
    \Xhline{2.\arrayrulewidth}
    \end{tabular}
    \end{minipage}
    \centering
    \begin{minipage}{0.35\textwidth}
    \centering
    \begin{overpic}[width=0.98\linewidth]{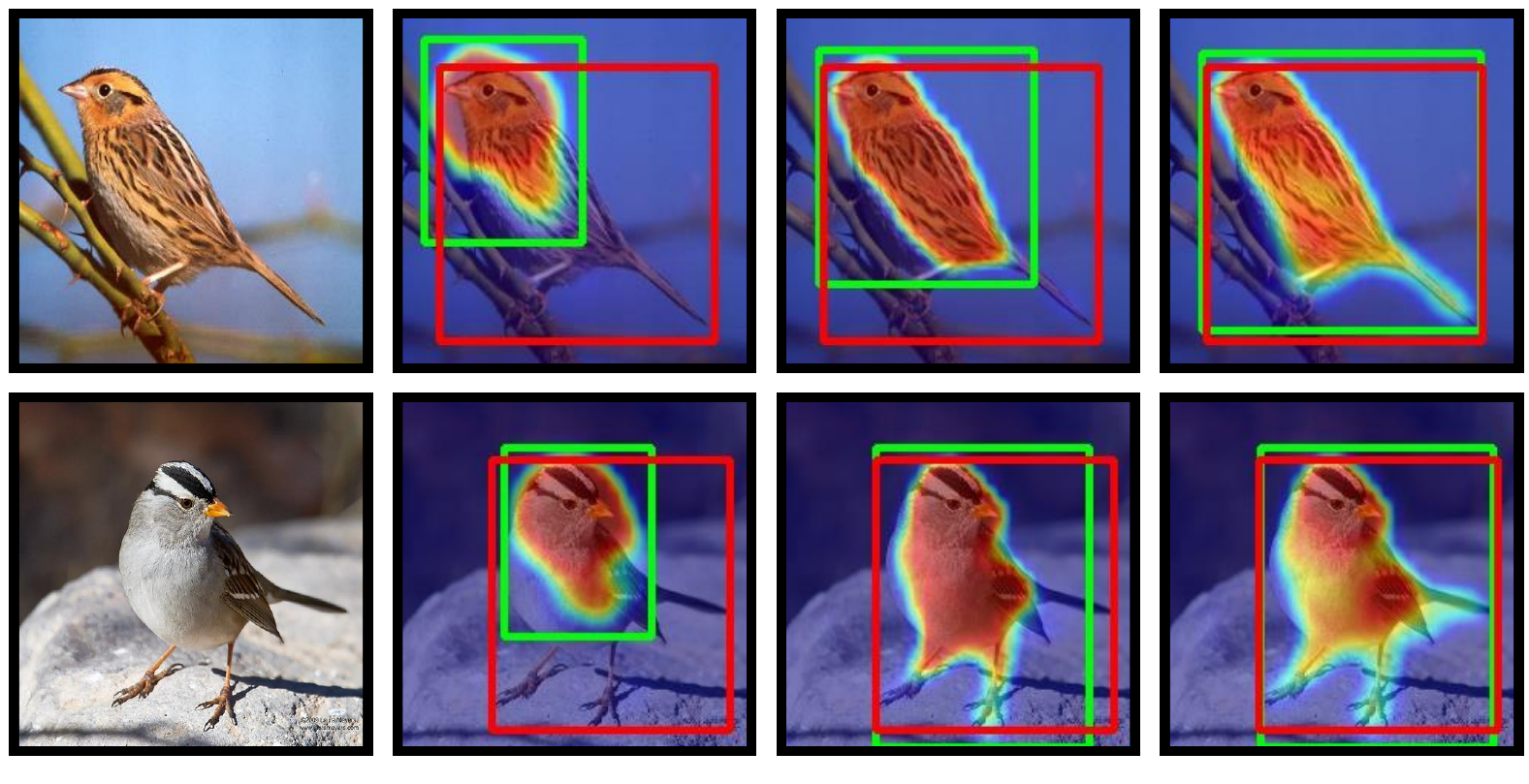}
	    \put(5, -2.5){\small\textbf{Image}}
	    \put(34, -2.5){\small\textbf{(a)}}
	    \put(59, -2.5){\small\textbf{(b)}}
	    \put(84.5, -2.5){\small\textbf{(c)}}
    \end{overpic}
    \end{minipage}
\end{table*}

\subsection{Comparison with State-of-the-Arts}
We compare the proposed BAS with state-of-the-art methods on \texttt{CUB-200-2011}~\citep{welinder2011caltech} and \texttt{ILSVRC}~\citep{russakovsky2015imagenet} datasets. As shown in Table~\ref{table:comparsion}, BAS achieves stable and excellent performance on various backbones. On \texttt{CUB-200-2011}~\citep{welinder2011caltech}, BAS surpasses all existing methods by a large margin in terms of GT-known/Top-1/Top-5 Loc when the backbones are MobileNetV1, ResNet50 and InceptionV3. Compared with the current Foreground-Prediction-Map-based method FAM~\citep{meng2021foreground}, BAS achieves \textbf{1.78\%}, \textbf{7.33\%}, \textbf{9.68\%} and \textbf{7.38\%} improvement on VGG16, MobileNetV1, ResNet50 and InceptionV3 in terms of GT-known Loc, respectively. On ResNet50, BAS achieves \textbf{95.41\%} GT-known Loc, which is a significant increase of \textbf{3.81\%} compared to the best performing counterpart Kim et al.~\citep{kim2022bridging}. In addition, our method improves \textbf{5.53\%} and \textbf{4.20\%} GT-known Loc compared to the latest multi-stage model CREAM~\citep{xu2022cream} on ResNet50 and InceptionV3, respectively.

\begin{figure*}[t]
\centering
	\begin{overpic}[width=0.99\linewidth]{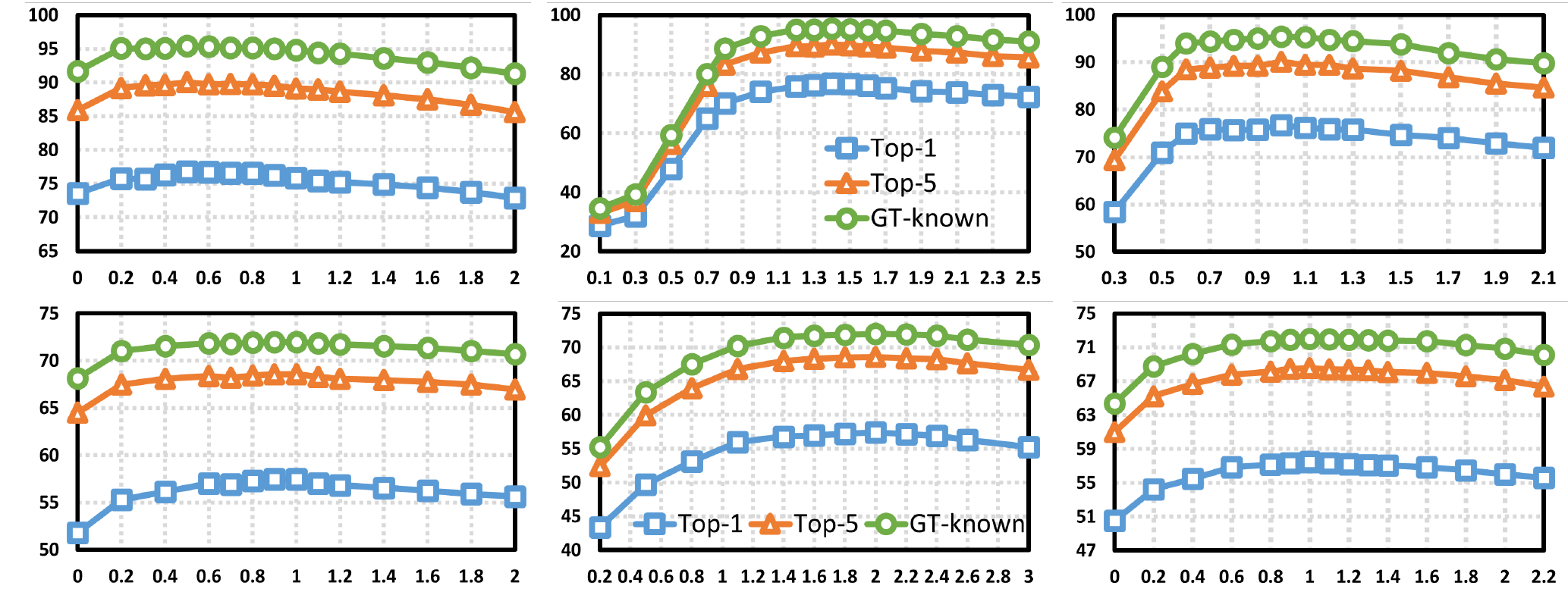}
	    \put(16.5, 0.){\small\textbf{(a)}}
	    \put(49.5, 0.){\small\textbf{(b)}}
	    \put(84., 0.){\small\textbf{(c)}}

        \put(0.4, 24.){\rotatebox{90}{\textbf{\texttt{CUB-200-2011}}}} 
        \put(0.4, 8.){\rotatebox{90}{\textbf{\texttt{ILSVRC}}}} 
\end{overpic}
\caption{\textbf{Hyperparameters.} \textbf{(a)} \bm{$\alpha$} for foreground region guidance loss $\mathcal{L}_{frg}$. \textbf{(b)} \bm{$\beta$} for area constraint loss $\mathcal{L}_{ac}$. \textbf{(c)} \bm{$\lambda$} for background activation suppression loss $\mathcal{L}_{bas}$.}
\label{hyper}
\end{figure*}

On \texttt{ILSVRC}~\citep{russakovsky2015imagenet}, BAS overall exceeds all baseline methods in terms of GT-known/Top-1/Top-5 Loc on all backbones. When MobileNetV1 is used as the backbone, our BAS achieves \textbf{72.03\%} GT-known Loc, surpassing FAM~\citep{meng2021foreground} by a large margin with a \textbf{9.98\%} improvement. Moreover, InceptionV3-BAS and ResNet50-BAS obtain \textbf{72.07\%} and \textbf{72.00\%} GT-known Loc, respectively, establishing a novel state-of-the-art. It shows that BAS performs well on both fine-grained dataset and large universal dataset. Furthermore, we visualize the localization maps of the proposed BAS and CAM~\citep{zhou2016learning} on \texttt{CUB-200-2011} and \texttt{ILSVRC} in Fig.~\ref{visual}. Compared to CAM, BAS can robustly cover the entire area of the object even in noisy environments and is sharper and more compact at the edges of the object.

\subsection{Ablation Study \label{ablation_section}}

In this section, we perform a series of ablation experiments using ResNet50~\citep{simonyan2014very} as the backbone. Above all, we conduct ablation experiments on various components of BAS on \texttt{CUB-200-2011}~\citep{welinder2011caltech}. We take $\mathcal{L}_{cls}$, $\mathcal{L}_{frg}$ and $\mathcal{L}_{ac}$ together as the baseline method for the Foreground-Prediction-Map-based architecture. As shown in Table~\ref{ablation}, the addition of $\mathcal{L}_{bas}$ to the baseline can enable the localization map to cover the object region more completely, thus significantly increasing the localization accuracy, with \textbf{21.01\%} and \textbf{16.26\%} improvement in terms of GT-known Loc and Top-1 Loc, respectively. Moreover, using Top-$k$ strategy to integrate the final localization result, though making the localization result not as sharp as before, it can further improve the GT-known Loc (from \textbf{92.15\%} to \textbf{95.41\%}) by alleviating the problem of the classification network focusing on the distinguish parts.

\begin{table*}[t]
    \centering
    \begin{minipage}{0.18\textwidth}
    \centering
    \caption{\textbf{Localization accuracy and visualization results} about inserting the generator after different layers on ResNet50.}
    \label{local_strategies}
    \end{minipage}
    \centering
    \begin{minipage}{0.36\textwidth}
    \renewcommand{\arraystretch}{1.}
    \renewcommand{\tabcolsep}{1.5pt}
    \small
    \centering
    \begin{tabular}{c|c|c|c}
\Xhline{2.\arrayrulewidth}
    \hline
       & \textbf{Top-1 Loc} & \textbf{Top-5 Loc} & \textbf{GT-known} \\
\Xhline{2.\arrayrulewidth}
    \hline
    \textbf{Layer 1} & 42.47  & 49.65     & 52.87    \\
    \textbf{Layer 2} & 69.24     & 81.24     & 86.16    \\
    \rowcolor{mygray}
    \textbf{Layer 3} & \textbf{76.75}     & \textbf{90.04}     & \textbf{95.41}    \\
    \textbf{Layer 4} & 71.63     & 84.81     & 90.94    \\
\Xhline{2.\arrayrulewidth}
    \hline
    \end{tabular}
    \end{minipage}
    \centering
    \begin{minipage}{0.43\textwidth}
    \centering
    \begin{overpic}[width=0.98\linewidth]{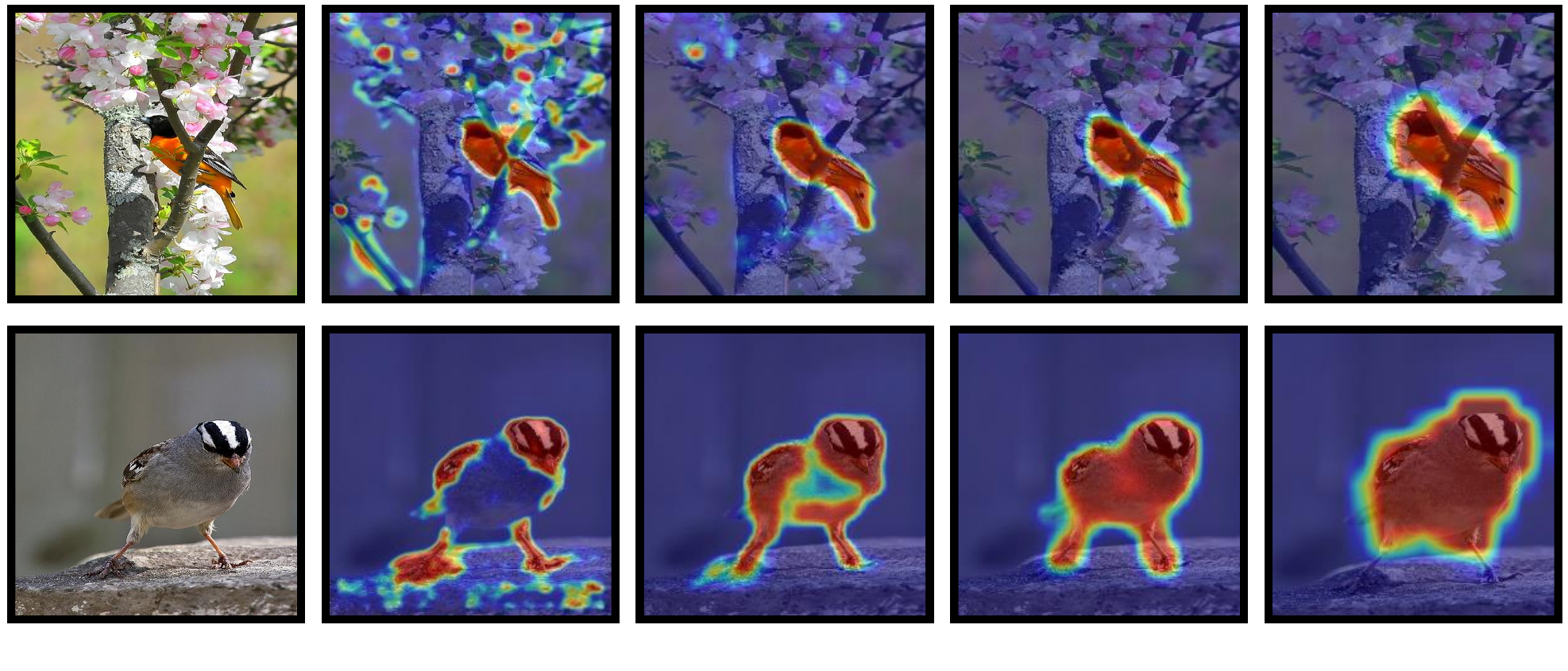}
	    \put(4 , -1){\small\textbf{Image}}
	    \put(23, -1){\small\textbf{layer 1}}
	    \put(43, -1){\small\textbf{layer 2}}
	    \put(63, -1){\small\textbf{layer 3}}
	    \put(83, -1){\small\textbf{layer 4}}
    \end{overpic}
    \end{minipage}
\end{table*}

\myPara{Hyperparameter \bm{$\alpha$}, \bm{$\beta$}, and \bm{$\lambda$} in total loss.} There are three hyperparameters in Equation~\ref{eq:WSOL}. Their effectiveness and sensitivity analyses for localization quality are performed on \texttt{CUB-200-2011} and \texttt{ILSVRC} in Fig.~\ref{hyper}. The $\alpha$ denotes the factor of $\mathcal{L}_{frg}$, and it can be noticed from Fig.~\ref{hyper} (a) that the presence of foreground region guidance loss ($\alpha$ $\ge$ 0.2) can significantly improve the localization accuracy by ensuring stable learning of foreground activation maps on both datasets. The $\beta$ reflects the degree of constraint between foreground area and background suppression. When $\beta$ is small, more areas in the foreground activation map are activated, while when $\beta$ is too large, it will suppress the learning of the activation map. As shown in Fig.~\ref{hyper} (b), our method performs stably with high accuracy when $\beta$ varies from 1.2 to 1.7 on \texttt{CUB-200-2011} and from 1.6 to 2.4 on \texttt{ILSVRC}. The $\lambda$ denotes the factor of $\mathcal{L}_{bas}$. A larger $\lambda$ indicates that more regions in the prediction map are activated by background activation suppression. As shown in Fig.~\ref{hyper} (c), the localization accuracy continues to grow on \texttt{CUB-200-2011} when $\lambda$ increases from 0.3 to 0.6 and remains stable from 0.6 to 1.3 with less than 1\% change in GT-known Loc, which shows that the proposed BAS approach can significantly improve the localization accuracy. In summary, although we have three hyperparameters in the loss function, it is easy to choose suitable values for the hyperparameters $\alpha$, $\beta$, and $\lambda$. In addition, we also provide the results of different combinations of hyperparameters on \texttt{CUB-200-2011} and \texttt{ILSVRC} in Section~\ref{ablation_section_WSSS}.

\begin{figure}[t]
\small
    \centering
    \begin{overpic}[width=1.\linewidth]{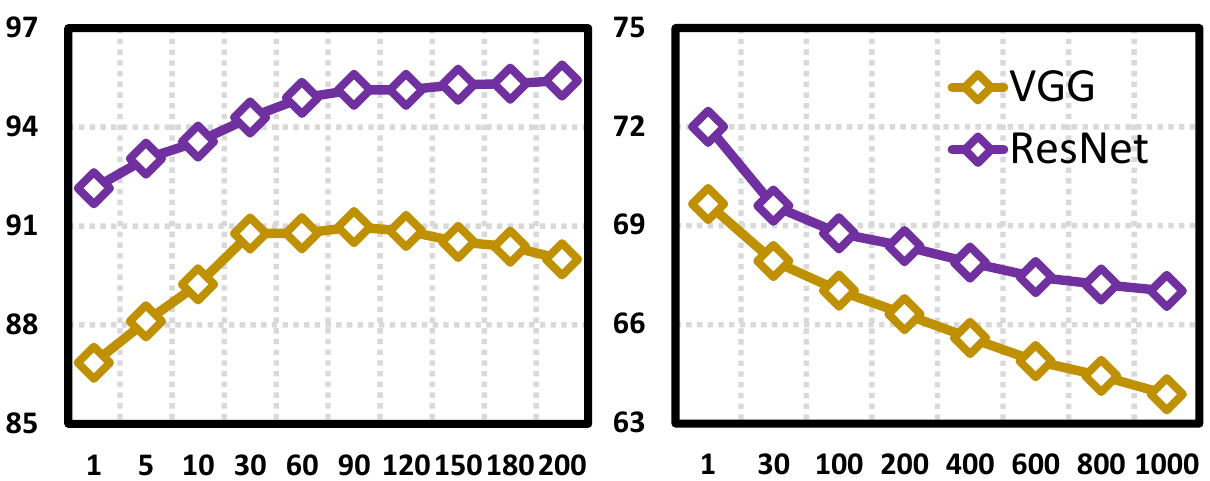}
	    \put(16, 39){\textbf{\texttt{CUB-200-2011}}}
	    \put(71.5, 39){\textbf{\texttt{ILSVRC}}}
    \end{overpic}
    \caption{\textbf{GT-known Loc ($\%$) \bm{$w.r.t$} \bm{$k$}.} Evaluation results of combining the Top-$k$ prediction maps when the backbone is VGG16 and ResNet50 respectively.}
    \label{top-k}
\end{figure}

\begin{figure}[t]
\centering
\small
    \begin{minipage}{0.20\textwidth}
    \renewcommand{\arraystretch}{1}
    \renewcommand{\tabcolsep}{1pt}
    \centering
        \begin{tabular}{c|c|c}
        \Xhline{2.\arrayrulewidth}
    \hline
               \textbf{Loc Acc}   & \textbf{(a)}  & \textbf{(b)} \\
        \Xhline{2.\arrayrulewidth}
    \hline
            \textbf{Top-1 Loc} & 74.97 & \cellcolor{mygray}\textbf{76.75} \\
            \textbf{Top-5 Loc} & 87.78 & \cellcolor{mygray}\textbf{90.04} \\
            \textbf{GT-known} & 93.75 & \cellcolor{mygray}\textbf{95.41} \\
        \Xhline{2.\arrayrulewidth}
    \hline
        \end{tabular}
    \end{minipage}
    \centering
    \begin{minipage}{0.265\textwidth}
    \centering
    \begin{overpic}[width=1.\linewidth]{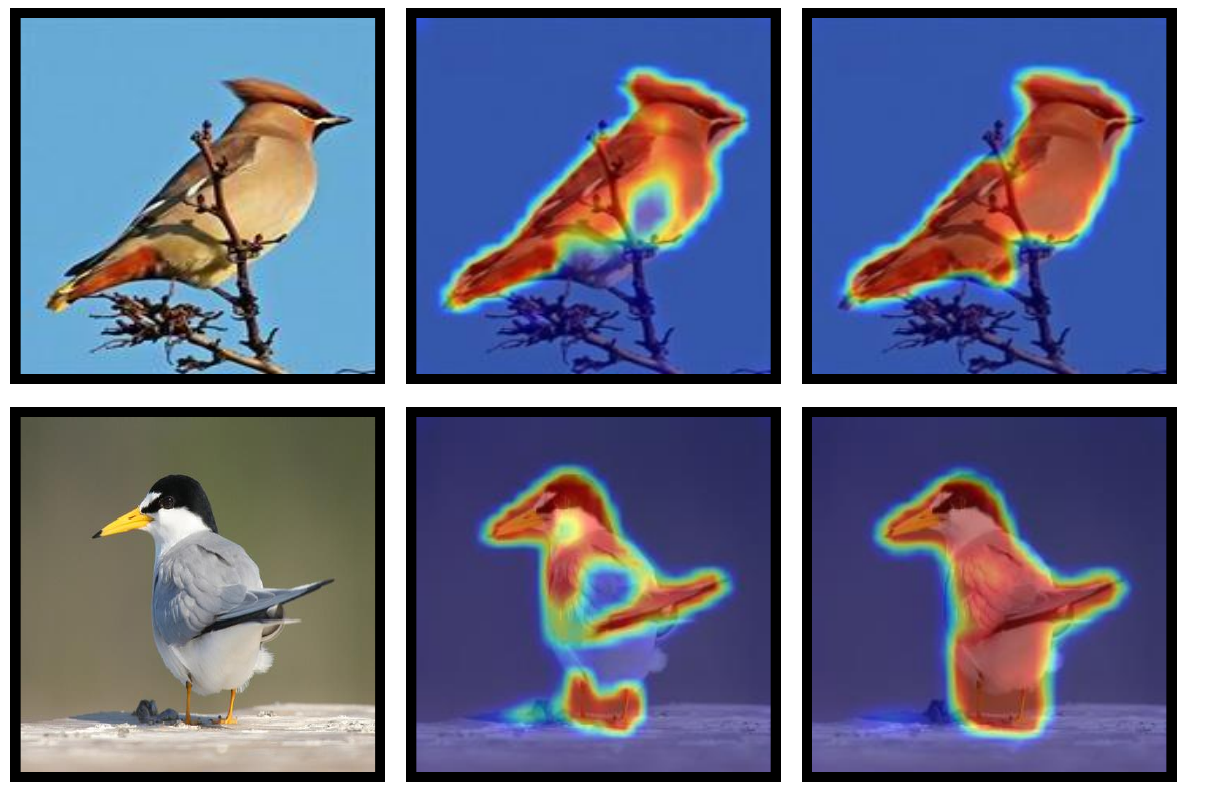}
	    \put(7, -4.7){\small\textbf{Image}}
	    \put(45, -4.7){\small\textbf{(a)}}
	    \put(76, -4.7){\small\textbf{(b)}}
    \end{overpic}
    \end{minipage}
    \caption{\textbf{Comparison of background prediction maps learned from (a) original image or (b) feature maps.}}
    \label{erase_strategies}
\end{figure}

\myPara{Hyperparameter \bm{$k$} in Top-\bm{$k$} strategy.} We evaluate the effect of the hyperparameter $k$ in our BAS. As shown in Fig.~\ref{top-k}, the accuracy of GT-known Loc is improved on \texttt{CUB-200-2011} when $k>1$, comparing $k=1$. For VGG16 and ResNet50, the highest localization accuracy is achieved at $k$ of 80 and 200, respectively. It suggests that the Top-$k$ strategy can be used to obtain more complete localization results and further improve the localization performance by integrating the localization maps of similar categories on \texttt{CUB-200-2011}. In contrast, for both VGG16 and ResNet50, the best localization results are obtained for $k=1$ on \texttt{ILSVRC} dataset, which shows a high variability of classes and few localization features of similarity between categories on \texttt{ILSVRC}.

\begin{table*}[t]
\renewcommand{\arraystretch}{1.}
\renewcommand{\tabcolsep}{5pt}
\small
\centering
\caption{\textbf{Evaluation results in terms of MaxBoxAccV2} on the \texttt{CUB-200-2011} and \texttt{ILSVRC} datasets using various backbones.}
\label{maxboxaccv2}
\begin{tabular}{l|c|cccc|cccc}
\Xhline{2.0\arrayrulewidth}
\hline
\multirow{2}{*}{\textbf{Methods}} & \multirow{2}{*}{\textbf{Venue}} & \multicolumn{4}{c|}{\textbf{\texttt{CUB-200-2011} (\textbf{MaxBoxAccV2})}} & \multicolumn{4}{c}{\textbf{\texttt{ILSVRC} (\textbf{MaxBoxAccV2})}} \\
\cline{3-10}
      &  & \textbf{VGG} & \textbf{Inception} & \textbf{ResNet} & \textbf{Mean} & \textbf{VGG} & \textbf{Inception} & \textbf{ResNet} & \textbf{Mean} \\
\hline
\Xhline{2.0\arrayrulewidth}
CAM~\citep{zhou2016learning} & CVPR$16$ & 63.7 & 56.7 & 63.0 & 61.1& 60.0 & 63.4 & 63.7 & 62.4 \\
    
    HaS~\citep{singh2017hide} & ICCV$17$ & 63.7 & 53.4 & 64.7 & 60.6  & 60.6 & 63.7 & 63.4 & 62.6\\
    
    ACoL~\citep{zhang2018adversarial} & CVPR$18$  & 57.4 & 56.2 & 66.5 & 60.0 & 57.4 & 63.7 & 62.3 & 61.1 \\
    
    SPG~\citep{zhang2018self} & ECCV$18$  & 56.3 & 55.9 & 60.4 & 57.5 & 59.9 & 63.3 & 63.3 & 62.2\\
    
    CutMix~\citep{yun2019cutmix} & ICCV$19$  & 62.3 & 57.5 & 62.8 & 60.8 & 59.4 & 63.9 & 63.3 & 62.2\\
    
    ADL~\citep{choe2020attention} & TPAMI$20$  & 66.3 & 58.8 & 58.3 & 61.1 & 59.8 & 61.4 & 63.7 & 61.7 \\
    
    IVR~\citep{kim2021normalization} & ICCV$21$  &  65.2 &  60.8 & 66.9 & 64.3 & 61.5 & 65.5 & 65.6 & 64.2\\
    
    DA-WSOL~\citep{zhu2022weakly} & CVPR$22$  & - & 68.0 & 69.9 & - & - & 64.8 & 68.2 & -\\
    
    CREAM~\citep{xu2022cream}  & CVPR$22$  &  71.5 & 64.2 & 73.5 & 69.7  & 66.2 & \textbf{68.9} & 67.4 & 67.5\\ 
    
    Kim et al.~\citep{kim2022bridging} & CVPR$22$  & 80.1 & - & 75.9 & - & 66.6 & - & 68.7& -\\
    
    C2AM~\citep{xie2022c2am} & CVPR$22$   &  81.4 & 82.4 & 83.8 & 82.5  & 66.3 & 65.8 & 66.8 & 66.3\\
    
    \hline
    \rowcolor{mygray}
    \textbf{BAS (ours)} & This Work  & \textbf{83.5} & \textbf{82.7} & \textbf{89.4} & \textbf{85.2} & \textbf{68.2} & \textbf{68.9} & \textbf{68.8} & \textbf{68.6}\\
\hline
\Xhline{2.0\arrayrulewidth}
\end{tabular}
\end{table*}

\myPara{Generator after different layers.} We report the results of inserting the generator after different layers of ResNet50. As shown in Table~\ref{local_strategies} (left table), quantitative experiment indicates that inserting the generator after \textbf{layer 3} achieves the best results and is significantly better than other positions. The prediction maps learned from different layers are visualized in Table~\ref{local_strategies} (right figure). When the generator learns localization information from shallow feature maps (\textbf{layer 1} and \textbf{layer 2}), the prediction map performs better at the edges of objects, but it is insufficient to resist background distractions and has poor semantic learning ability. In addition, the generator learns localization information from the high-level feature (\textbf{layer 4}) resulting in imprecise localization due to the limitation of resolution. 

\myPara{Original image $vs$ feature maps.} We fix the generator after layer 3 and conduct experiments on the masking position (original image $vs$ feature maps) of the background prediction map. As illustrated in Fig.~\ref{erase_strategies}, it can be noted that the masking feature maps approach achieves higher accuracy and better coverage of the localization results on the object, while the results generated by the approach of masking the original image focus more on the edge or texture of the object and have less ability to locate smooth regions. It may be because the learning process in shallow layers usually focuses on common basic features (\eg, edges, textures) and ignores high-level semantic features.

\begin{table}[t]
    \small
\centering
\caption{\textbf{Improvement in GT-known Loc compared to the previous conference version.}}
        \label{table:version_improvement}
\renewcommand{\arraystretch}{1}
\renewcommand{\tabcolsep}{12pt}
\begin{tabular}{c|c|c}
\Xhline{2.\arrayrulewidth}
\hline 
\textbf{Backbone} & \texttt{CUB-200-2011} & \texttt{ILSVRC} \\
\Xhline{2.\arrayrulewidth} 
\hline 
VGG16 & 91.04 \color{gray}({$-$0.03}) & 69.66 \color{red}(\textbf{+0.02})\\ 
MobileNetV1 & 93.04 \color{red}(\textbf{+0.69}) & 72.03 \color{red}(\textbf{+0.03})\\ 
ResNet50 & 95.41 \color{red}(\textbf{+0.28}) & 72.00 \color{red}(\textbf{+0.23})\\ 
InceptionV3 & 94.63 \color{red}(\textbf{+2.39}) & 72.07 \color{red}(\textbf{+0.14}) \\

\hline 
\rowcolor{mygray}
\textbf{Mean} & 93.53 \color{red}(\textbf{+0.83}) & 71.44 \color{red}(\textbf{+0.11}) \\

\hline
\Xhline{2.\arrayrulewidth}
\end{tabular}
\end{table}

\begin{figure}[t]
\small
    \centering
    \begin{overpic}[width=0.98\linewidth]{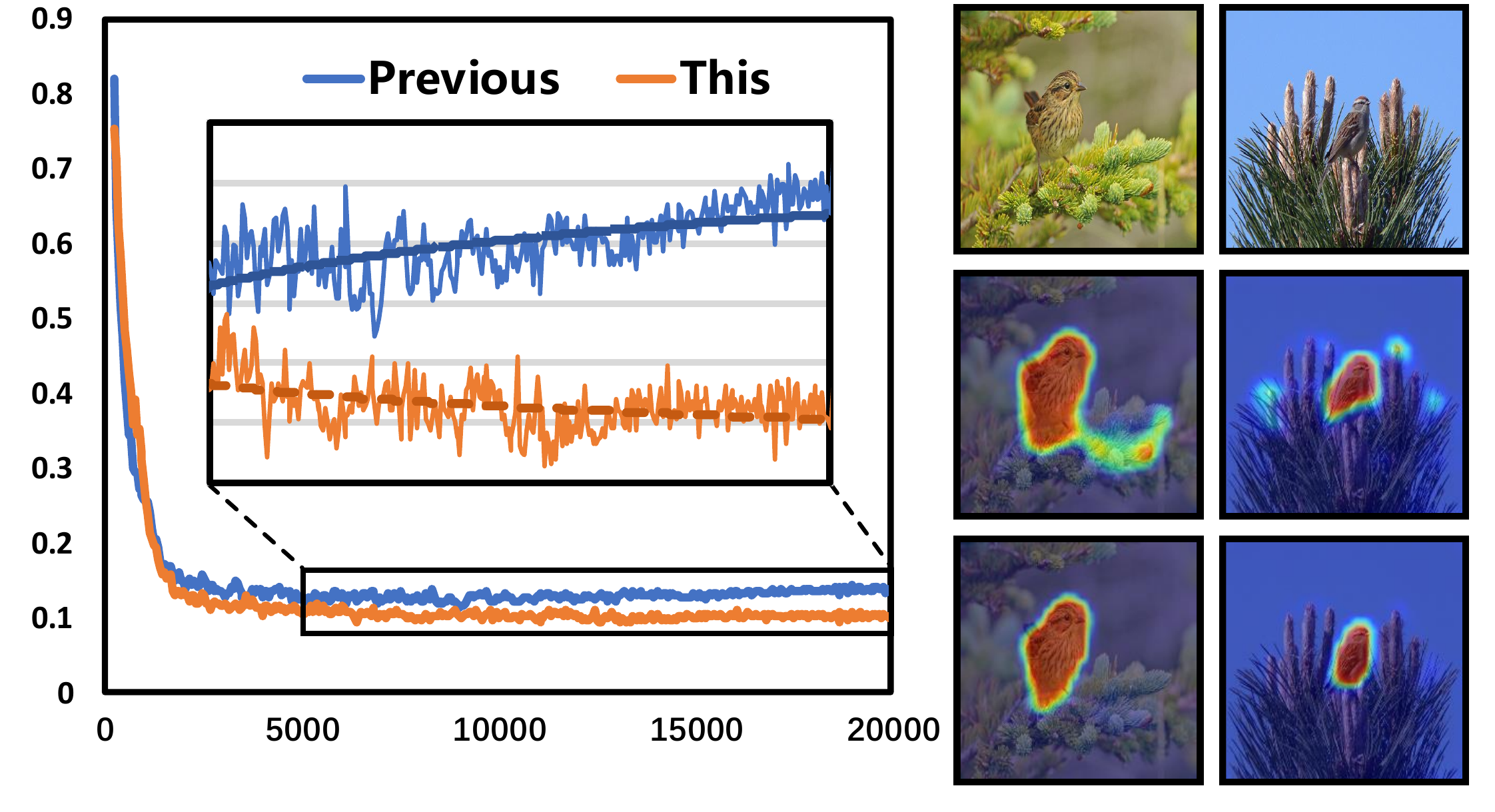}
    \put(30.5, -1.3){\small\textbf{(a)}}
    \put(77.8, -1.3){\small\textbf{(b)}}
    \put(24.5, 2.){\textbf{Iterations}}
    \put(-1.5, 23.5){\rotatebox{90}{\textbf{BAS Loss}}}
    \put(99, 51.5){\rotatebox{270}{\textbf{Image}}}
    \put(99, 36.2){\rotatebox{270}{\textbf{Previous}}}
    \put(99, 15.){\rotatebox{270}{\textbf{This}}}
    \end{overpic}
    \caption{\textbf{Experimental comparison between the previous conference version and this work.} \textbf{(a)} The $\mathcal{L}_{bas}$ training loss curves. \textbf{(b)} Visualization of the localization results.}
    \label{fig:bas_loss}
\end{figure}
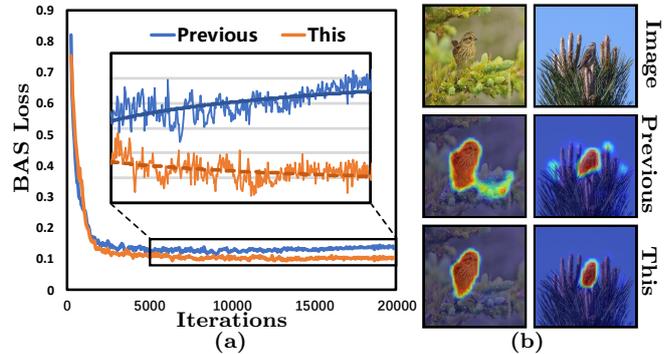
\myPara{Comparison with previous conference version.} We compare the improvement over the conference version in both quantitative and qualitative aspects. Benefiting from the adjustment to the position of the \texttt{ReLU} activation function, BAS can learn the feature map more adequately and efficiently. As shown in Fig.~\ref{fig:bas_loss} (a), we display the curve of $\mathcal{L}_{bas}$ (Equation \ref{eq:BAS}) training loss with the training iterations for both previous conference version and this work. It can be observed that the loss curve (this work) converges to a lower point and shows a more stable convergence trend, while the loss curve in the previous conference version even presents an increasing trend during the iterations. It indicates that the \texttt{ReLU} in the last layer (Fig.~\ref{fig:relu}) makes the classification network learn the background region insufficiently, hence resulting in the inadequate convergence of BAS loss. Fig.~\ref{fig:bas_loss} (b) illustrates some localization maps to support this analysis. Compared with the previous conference version, BAS (this work) demonstrates more robustness in the learning of the background region and consequently improves the localization accuracy in Table~\ref{table:version_improvement}. We achieve an average of \textbf{0.83\%} and \textbf{0.11\%} GT-known Loc gains on the four backbone networks on \texttt{CUB-200-2011} and \texttt{ILSVRC}, respectively, without additional parameters and computations.

\subsection{Performance Analysis}

In this section, we evaluate and analyze in detail the localization quality and segmentation quality of BAS.

\begin{figure}[t]
\small
    \centering
    \begin{overpic}[width=1.\linewidth]{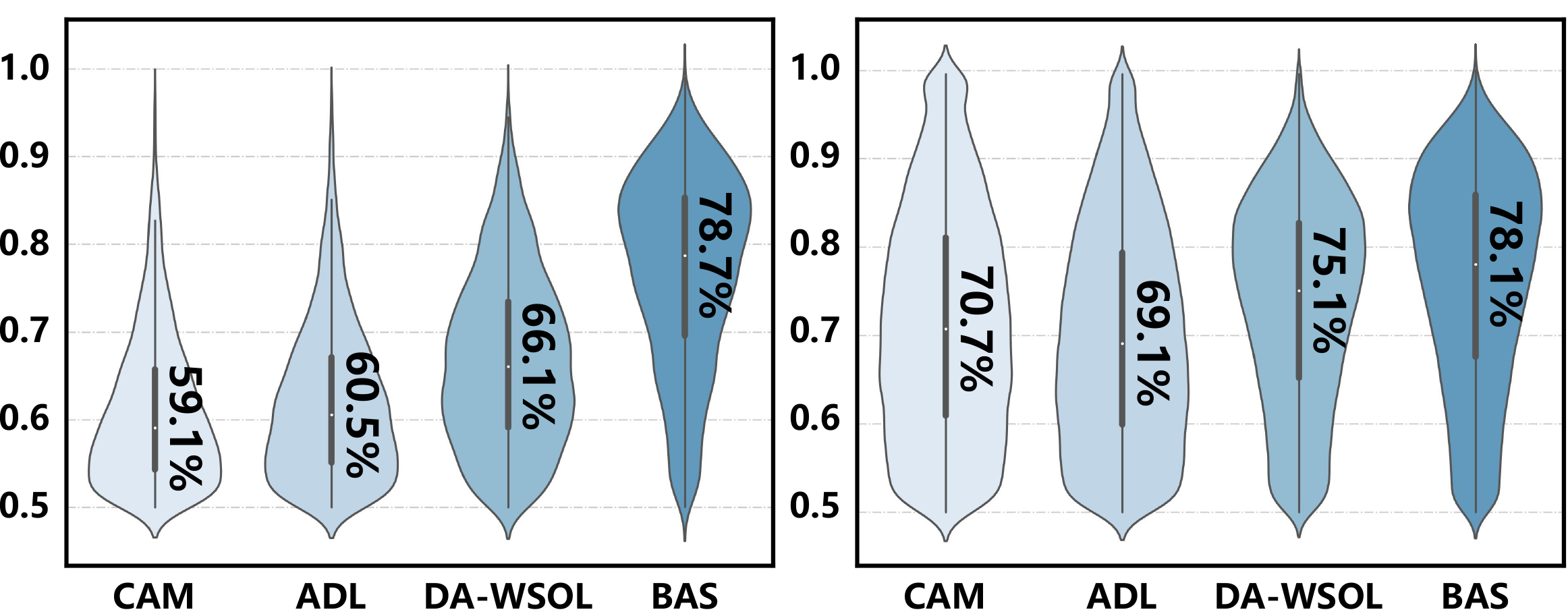}
	    \put(16, 39.5){\textbf{\texttt{CUB-200-2011}}}
	    \put(71, 39.5){\textbf{\texttt{ILSVRC}}}
    \end{overpic}
    \caption{\textbf{Statistical analysis of correct bounding boxes}, based on ResNet50 (CAM~\citep{zhou2016learning}, ADL~\citep{choe2020attention}, and DA-WSOL~\citep{zhu2022weakly}).}
    \label{localization_iou}
\end{figure}
\begin{figure*}[t]
\centering
\small
	\begin{overpic}[width=1.0\linewidth]{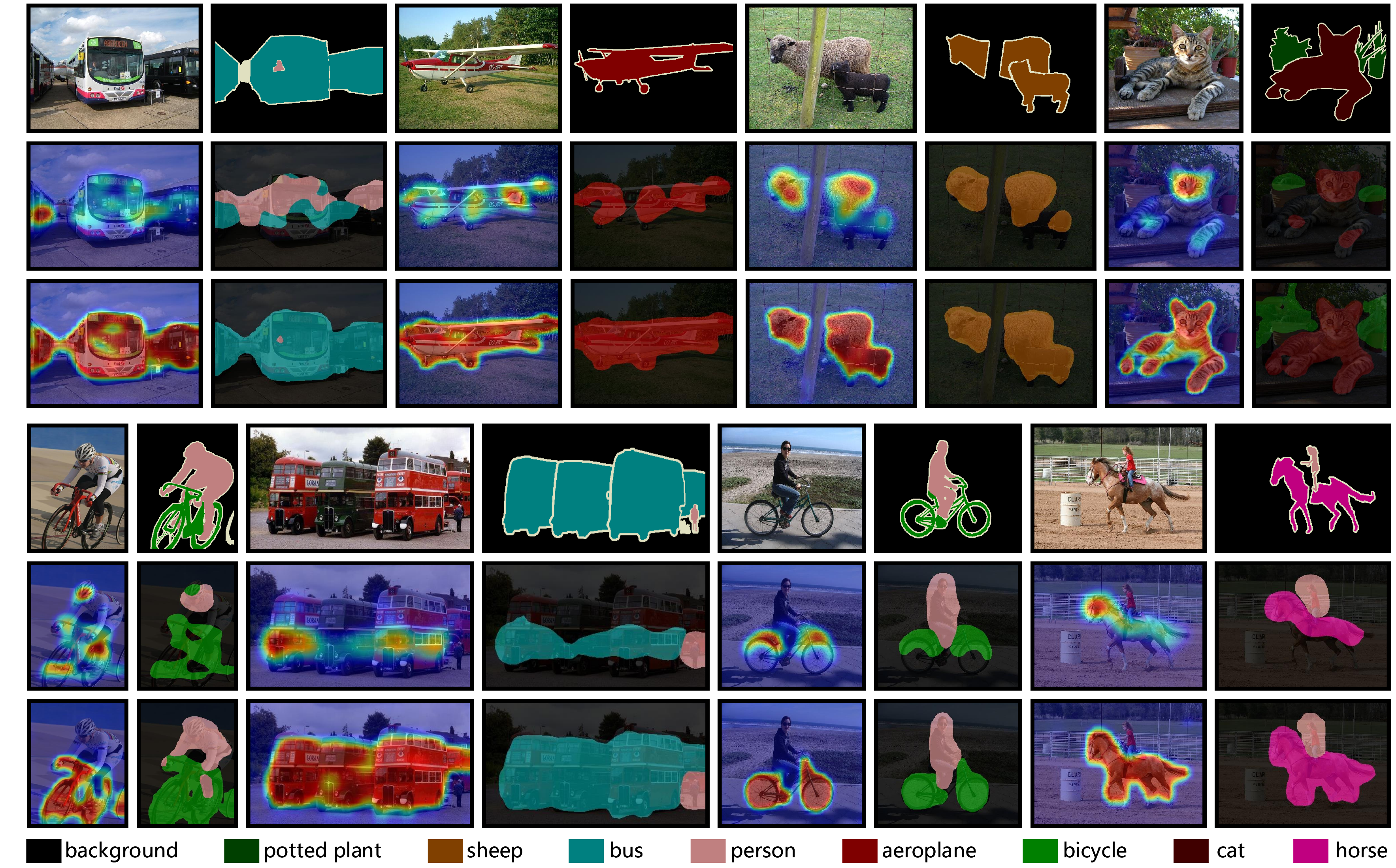}
	    \put(0.3, 5.6){\rotatebox{90}{\small\textbf{Ours}}}
	    \put(0.3, 15.0){\rotatebox{90}{\small\textbf{CAM}}}
	    \put(0.3, 26.){\rotatebox{90}{\small\textbf{GT}}}
	    
	    \put(0.3, 35.4){\rotatebox{90}{\small\textbf{Ours}}}
	    \put(0.3, 45.2){\rotatebox{90}{\small\textbf{CAM}}}
	    \put(0.3, 56.0){\rotatebox{90}{\small\textbf{GT}}}
\end{overpic}
\caption{\textbf{Visualization of the initial seed generated by CAM and the proposed BAS} on the \texttt{PASCAL VOC 2012} dataset.}
\label{seed_voc}
\end{figure*}

\begin{figure*}[t]
\footnotesize
\centering
\begin{minipage}{0.44\linewidth}
    \renewcommand{\arraystretch}{1}
    \renewcommand{\tabcolsep}{2.pt}
    \small
    \centering
    \begin{tabular}{l|c|cc|cc}
    \Xhline{2.\arrayrulewidth}
    \hline
    \multirow{2}{*}{\textbf{Method}}& \multirow{2}{*}{\textbf{Venue}}& \multicolumn{2}{c|}{\textbf{\texttt{CUB-200-2011}}} & \multicolumn{2}{c}{\textbf{\texttt{OpenImages}}} \\
    \cline{3-6}
    & & \textbf{PIoU} & \textbf{PxAP} & \textbf{PIoU} & \textbf{PxAP}\\
    \Xhline{2.\arrayrulewidth}
    \hline
    CAM & CVPR$16$   & 42.52 & 63.70 & 42.74 & 57.54  \\
    ADL & TPAMI$20$  & 41.51 & 58.17 & 41.59 & 54.39 \\
    DA-WSOL & CVPR$22$  &  56.18 & 74.70 & 49.68 & 65.42 \\
    \hline
    \rowcolor{mygray}
    \textbf{Ours}  & This Work  & \textbf{71.24} & \textbf{89.94} & \textbf{50.72} & \textbf{66.86}   \\
    \hline
    \Xhline{2.\arrayrulewidth}
    \end{tabular}
\end{minipage}
\centering
\begin{minipage}{0.55\linewidth}
    \centering
    \begin{overpic}[width=1.\linewidth]{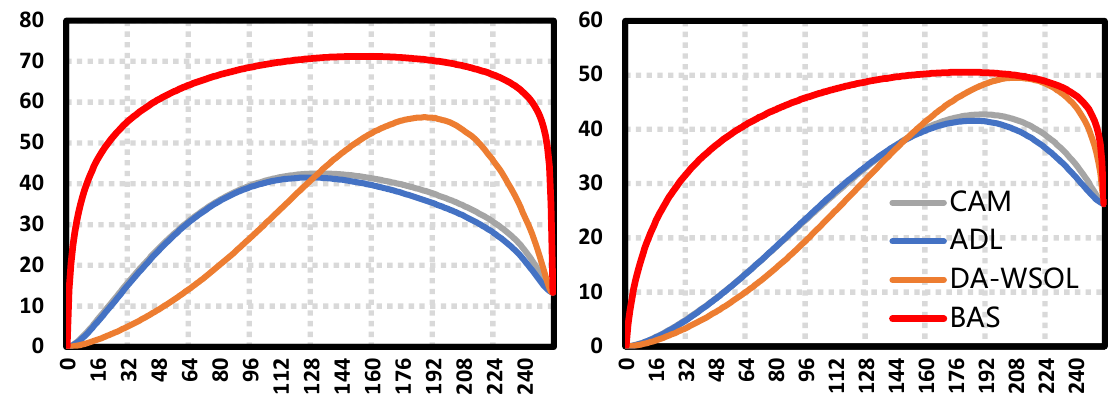}
	    \put(18.5, 35.5){\textbf{\texttt{CUB-200-2011}}}
	    \put(70, 35.5){\textbf{\texttt{OpenImages}}}
	    \put(-1,17.5){\rotatebox{90}{\textbf{IoU}}}
    \end{overpic}
\end{minipage}
    \caption{\textbf{Segmentation Quality.} IoU-Threshold curves for different baseline methods and evaluation results of PIoU, PxAP on \texttt{CUB-200-2011}~\citep{welinder2011caltech} and \texttt{OpenImages} \citep{choe2020evaluating} datasets, based on ResNet50 (CAM~\citep{zhou2016learning}, ADL\citep{choe2020attention}, and DA-WSOL~\citep{zhu2022weakly}).}
    \label{segmentation_iou}
\end{figure*}

\begin{figure*}[t]
\centering
	\begin{overpic}[width=1.\linewidth]{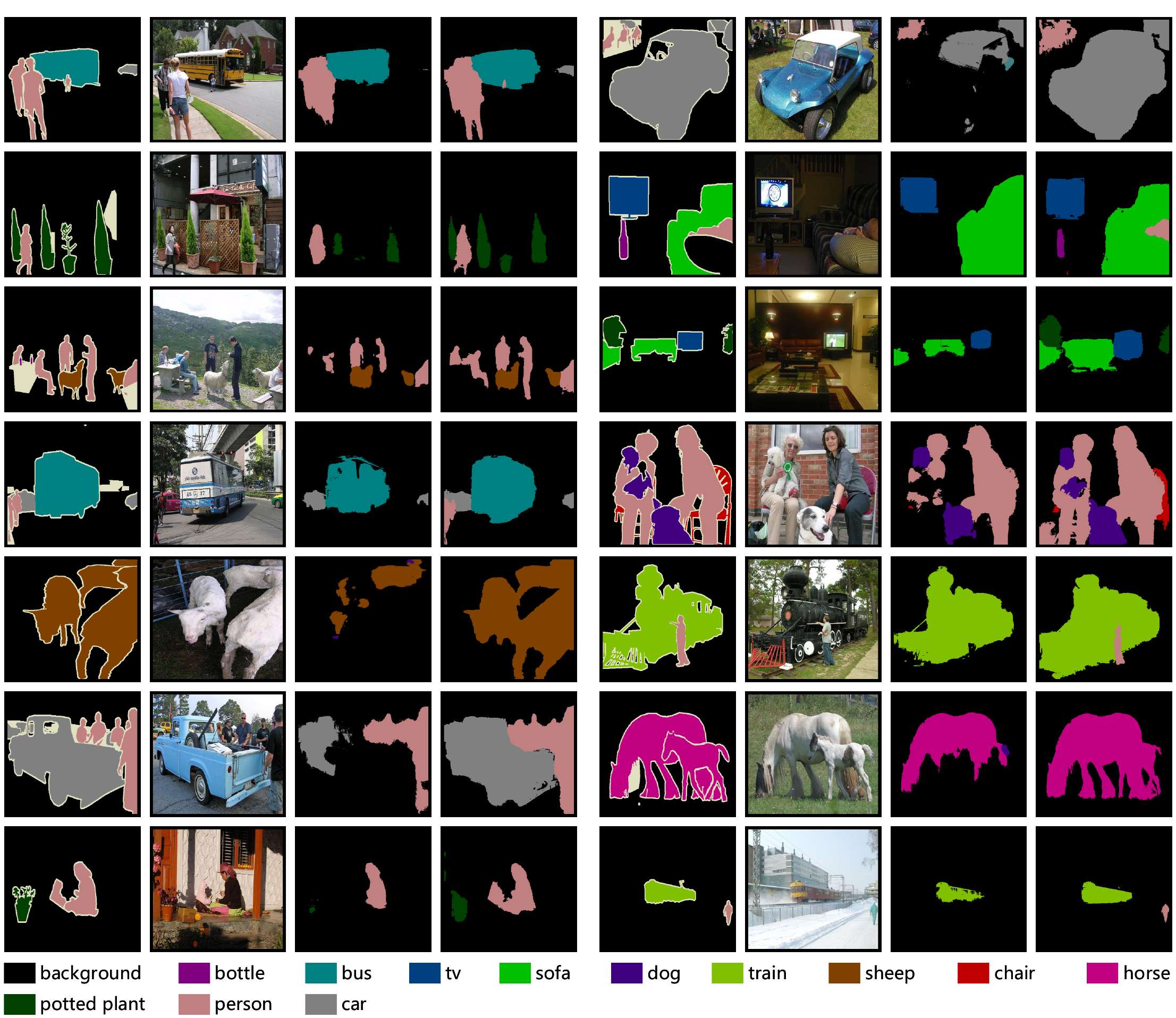}
	    \put(5, 85.5){{\small\textbf{GT}}}
	    \put(16, 85.5){{\small\textbf{Image}}}
	    \put(29., 85.5){{\small\textbf{IRN}}}
	    \put(41, 85.5){{\small\textbf{BAS}}}
	    
	    \put(55.0, 85.5){{\small\textbf{GT}}}
	    \put(66.8, 85.5){{\small\textbf{Image}}}
	    \put(79.8, 85.5){{\small\textbf{IRN}}}
	    \put(91.8, 85.5){{\small\textbf{BAS}}}

\end{overpic}
\caption{\textbf{Examples of semantic segmentation results} on \texttt{PASCAL VOC 2012} for IRN and BAS (with IRN).}
\label{mask_voc}
\end{figure*}

\begin{figure*}[t]
\centering
\small
	\begin{overpic}[width=1.0\linewidth]{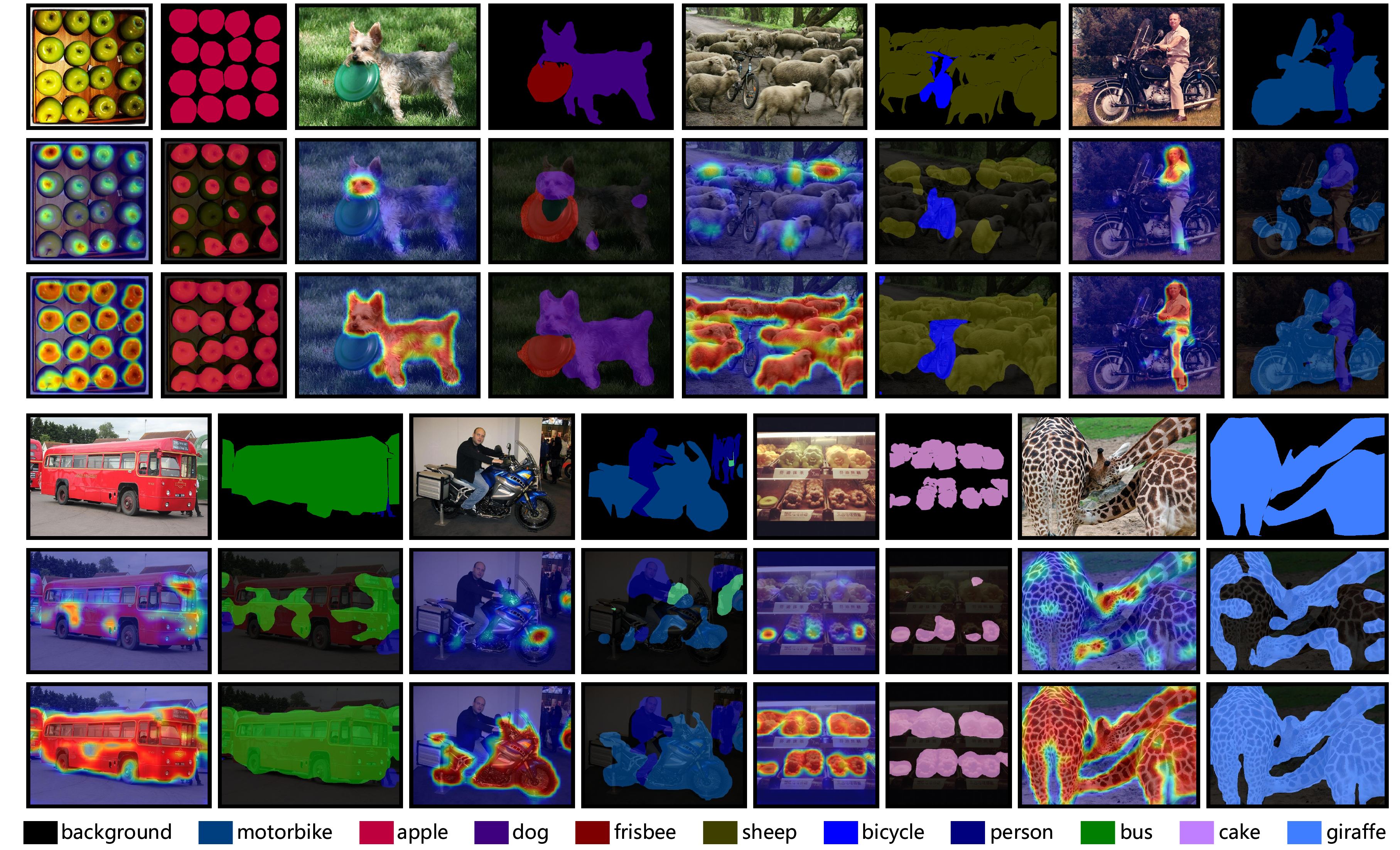}
	    \put(0.3, 5.){\rotatebox{90}{\small\textbf{Ours}}}
	    \put(0.3, 14.4){\rotatebox{90}{\small\textbf{CAM}}}
	    \put(0.3, 25.1){\rotatebox{90}{\small\textbf{GT}}}
	    
	    \put(0.3, 34.5){\rotatebox{90}{\small\textbf{Ours}}}
	    \put(0.3, 44.){\rotatebox{90}{\small\textbf{CAM}}}
	    \put(0.3, 54.8){\rotatebox{90}{\small\textbf{GT}}}
\end{overpic}
\caption{\textbf{Visualization of the initial seed generated by CAM and the proposed BAS} on the \texttt{MC COCO 2012} dataset.}
\label{seed_coco}
\end{figure*}
\myPara{Localization Quality.} Tabel~\ref{maxboxaccv2} shows the MaxBoxAccv2 scores compared with other methods on \texttt{CUB-200-2011}~\citep{welinder2011caltech} and \texttt{ILSVRC}~\citep{russakovsky2015imagenet}. Quantitative experiments indicate that our method achieves the best results for different backbone networks and datasets under the MaxBoxAccv2 criterion, which proves the high quality of the bounding box generated by BAS and verifies the effectiveness and generalizability of the proposed method. In particular, on \texttt{CUB-200-2011}, we exceed the previous best methods by 2.1\% and 5.6\% when the backbone networks are VGG16 and ResNet50, respectively. Besides, in Fig.~\ref{localization_iou}, we demonstrate the statistical analysis of IoU based on ResNet50, which plots the IoU distribution curves between the bounding boxes and the ground-truth boxes when localized correctly, following DANet~\citep{xue2019danet}. On \texttt{CUB-200-2011}, we achieve \textbf{78.7\%} IoU median, exceeding the latest state-of-the-art method DA-WSOL~\citep{zhu2022weakly} by \textbf{12.6\%}, and correspondingly by \textbf{3.0\%} on \texttt{ILSVRC}. From the median IoU and the IoU distribution, it can be seen that the proposed BAS significantly improves the localization quality on both \texttt{CUB-200-2011} and \texttt{ILSVRC} datasets.

\myPara{Segmentation Quality.} We compare the localization map with the ground-truth mask label using two metrics, PIoU and PxAP, following DA-WSOL~\citep{zhu2022weakly}. As shown in Fig.~\ref{segmentation_iou} (left table), we evaluate the performance of the proposed BAS with CAM~\citep{zhou2016learning}, ADL~\citep{choe2020attention} and DA-WSOL~\citep{zhu2022weakly} on ResNet50. Compared to DA-WSOL, BAS achieves significant and consistent improvement, with a \textbf{15.06\%} increase in PIoU and \textbf{15.24\%} in PxAP on \texttt{CUB-200-2011}. The proposed method also surpasses all methods on \texttt{OpenImages}, although \texttt{OpenImages} is a more challenging dataset due to a large number of small objects and complex backgrounds. In addition, we present the IoU-Threshold curves in the right graph of Fig.~\ref{segmentation_iou}, which represent the IoU values at varying thresholds within the range of [0, 255]. As observed from the IoU-Threshold curves on both datasets, our method demonstrates a lower sensitivity to the thresholds and achieves better results at arbitrary threshold compared to other methods, which indicates that the localization map produced by BAS has fewer low confidence regions and is closer to the ground-truth object region.

\section{Experiments on Weakly Supervised Semantic Segmentation \label{sec:WSSS_experiment}}

\subsection{Experimental Setup}

\myPara{Datasets and Evaluation Metric.} To evaluate the performance of BAS on weakly supervised semantic segmentation task, we conduct experiments on the commonly used \textbf{\texttt{PASCAL VOC 2012}}~\citep{everingham2010pascal} and \textbf{\texttt{MS COCO 2014}}~\citep{lin2014microsoft} datasets. \texttt{PASCAL VOC 2012} contains 21 categories (including one background class). It has 1,464, 1,449, and 1,456 samples in training, val, and test sets, respectively. Following the common experimental protocol~\citep{chen2014semantic}, the training set is augmented with 10,582 weakly annotated images provided by SBD dataset~\citep{hariharan2011semantic}.
\texttt{MS COCO 2014} dataset has 81 semantic classes (including one background class). Following~\cite{lee2022anti,jiang2022l2g}, images without the target categories are moved off the dataset, remaining 82,081 training images and 40,137 validation images. We use the mean Intersection-over-Union (\textbf{mIoU}) as the evaluation metric for all experiments.

\myPara{Implementation Details.} For seed generation, the input image is resized to 512$\times$512, then augmented by horizontal flipping and random cropping to 448$\times$448. We train the network for 10 epochs. Batch size is set to 16 and 64 on \texttt{PASCAL VOC 2012} and \texttt{MS COCO 2014} respectively. To optimize the network, SGD optimizer is adopted with momentum mechanism and the momentum coefficient is set to 0.9. The initial learning rate is set as 0.005 and decayed following the poly policy $lr_{\text{init}}$ = $lr_{\text{init}}(1-itr/max\_itr)^\rho$ with $\rho$ = 0.9. Following~\cite{lee2022anti,xie2022clims}, we use ResNet50 as the backbone network to generate the initial seed for both \texttt{PASCAL VOC 2012} and \texttt{MS COCO 2014} datasets.

\myPara{Seed Refinement and Segmentation.} For seed refinement, to make a fair comparison, we follow~\cite{lee2022anti,lee2021reducing,chen2022class} using IRN~\citep{ahn2019weakly} to improve the quality of the initial seed. After generating pseudo masks, we select DeepLabV2~\citep{chen2017deeplab} with ResNet-101~\citep{he2016deep} as the segmentation network, following~\cite{xie2022clims,jo2021puzzle}. We adopt the default setting to train DeepLabV2 as in~\cite{lee2022anti} with weights pretrained on \texttt{MS COCO 2014}.

\begin{table}[t]
    \small
\centering 
\caption{\textbf{Ablation study} for the components of BAS on \texttt{PASCAL VOC 2012} and \texttt{MS COCO 2014}.}
\label{table:ablation_WSSS}
\renewcommand{\arraystretch}{1}
\renewcommand{\tabcolsep}{16pt}
\begin{tabular}{cc|cc}
\Xhline{2.\arrayrulewidth}
\hline 
\textbf{Baseline} & \bm{$\mathcal{L}_{bas}$} & \texttt{\textbf{PASCAL}} & \texttt{\textbf{COCO}}\\
\Xhline{2.\arrayrulewidth}
\hline  
\checkmark  & & 50.1 & 32.5\\   
\hline 
\rowcolor{mygray}
\checkmark  &\checkmark   & \textbf{57.7} & \textbf{36.9}\\   
\hline  
\Xhline{2.\arrayrulewidth}
\end{tabular}
\end{table}

\subsection{Ablation Study \label{ablation_section_WSSS}}

In this section, we perform a series of ablation experiments with ResNet50 as the backbone on \texttt{PASCAL VOC 2012} and \texttt{MS COCO 2014}. We first execute an ablation study regarding the loss composition of the BAS, and as in Section~\ref{ablation_section}, we take $\mathcal{L}_{cls}$, $\mathcal{L}_{frg}$, and $\mathcal{L}_{ac}$ together as the baseline for the Foreground-Prediction-Map-based architecture. It can be seen from Table~\ref{table:ablation_WSSS} that the addition of $\mathcal{L}_{bas}$ can significantly improve the segmentation quality of baseline with \textbf{7.6\%} and \textbf{4.4\%} mIoU gains on \texttt{PASCAL VOC 2012} and \texttt{MS COCO 2014}, respectively, which verifies the effectiveness of the proposed $\mathcal{L}_{bas}$ in capturing object regions relevant to classification.

\begin{figure}[t]
\centering
	\begin{overpic}[width=0.99\linewidth]{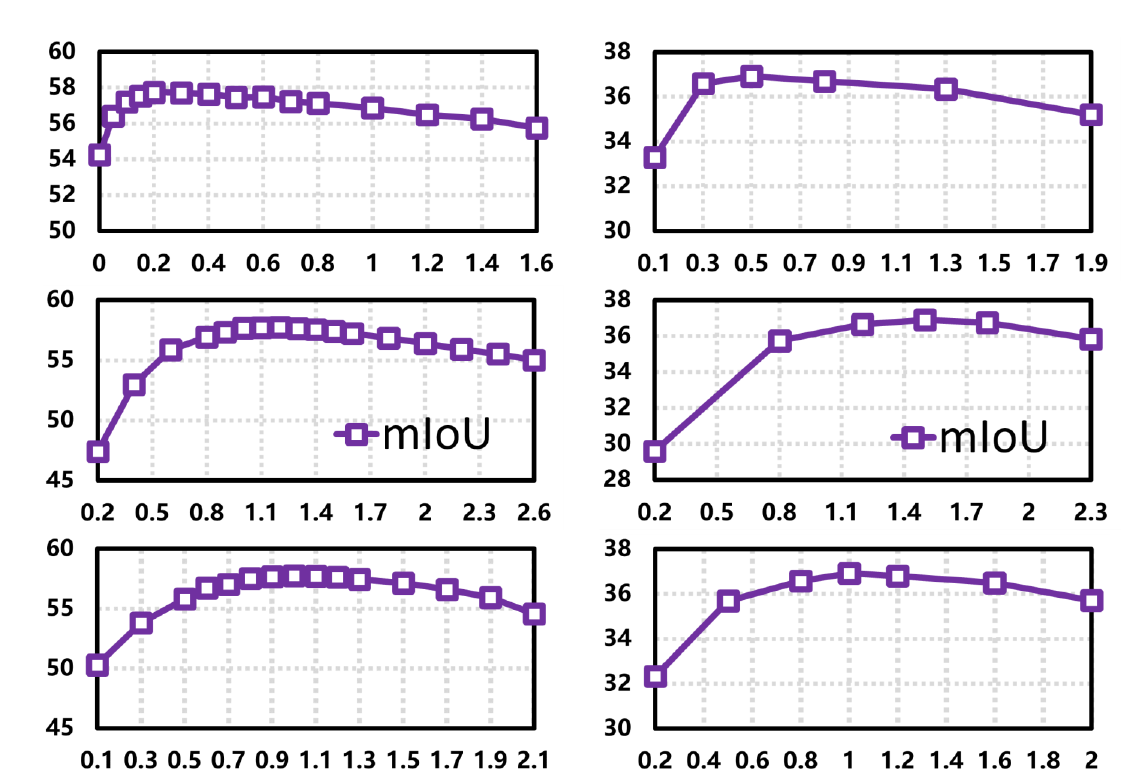}
	    \put(11.5, 65.8){\textbf{\texttt{PASCAL VOC 2012}}}
	    \put(63.5, 65.8){\textbf{\texttt{MS COCO 2014}}}

        \put(0., 10.){\rotatebox{90}{\small\textbf{(c)}}}
        \put(0., 32.){\rotatebox{90}{\small\textbf{(b)}}}
        \put(0., 54.){\rotatebox{90}{\small\textbf{(a)}}}
\end{overpic}
\caption{\textbf{Hyperparameters.} \textbf{(a)} \bm{$\alpha$} for foreground region guidance loss $\mathcal{L}_{frg}$. \textbf{(b)} \bm{$\beta$} for area constraint loss $\mathcal{L}_{ac}$. \textbf{(c)} \bm{$\lambda$} for background activation suppression loss $\mathcal{L}_{bas}$.}
\label{hyper_WSSS}
\end{figure}
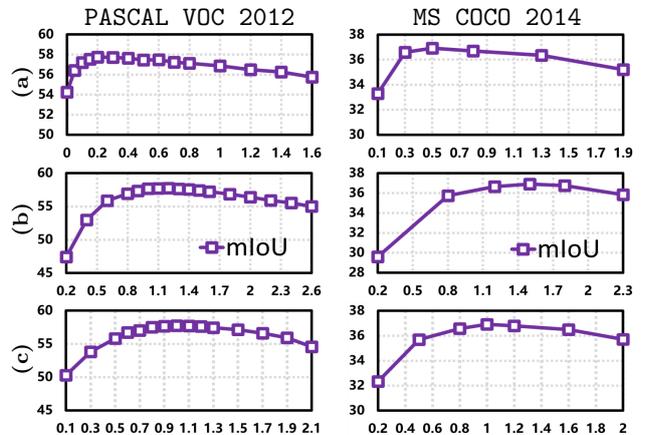

\myPara{Hyperparameter \bm{$\alpha$}, \bm{$\beta$}, and \bm{$\lambda$} in total loss.} Fig.~\ref{hyper_WSSS} illustrates the sensitivity of the segmentation quality to the hyperparameters $\alpha$, $\beta$, $\lambda$ on \texttt{PASCAL VOC 2012} and \texttt{MS COCO 2014}. Among them, $\alpha$ is the coefficient of $\mathcal{L}_{frg}$ and a small $\alpha$ can enable $\mathcal{L}_{frg}$ to work well. As shown in Fig.~\ref{hyper_WSSS} (a), the mIoU result is significantly improved on \texttt{PASCAL VOC 2012} when the $\alpha$ is greater than 0.1 and varies very little in the interval 0.15 to 0.5 with less than 0.3\% mIoU change. $\mathcal{L}_{ac}$ aims to constrain the foreground area to avoid unlimited expansion of the foreground area. Therefore, if the coefficient $\beta$ of $\mathcal{L}_{ac}$ is too small, it will lead to too many regions to be activated hence drastically reducing the segmentation performance as shown in Fig.~\ref{hyper_WSSS} (b). The purpose of $\mathcal{L}_{bas}$ is to allow the localization map to learn regions contributing to the classification in a background activation suppression manner. As shown in Fig.~\ref{hyper_WSSS} (c), the mIoU result remains stable on both datasets when the factor $\lambda$ of $\mathcal{L}_{bas}$ is in the range of 0.8 to 1.2.

\begin{table}[t]
    \small
\centering 
\caption{\textbf{Effect of different combinations of hyperparameters} on WSOL and WSSS with ResNet50 backbone. Report: the result reported in the paper. OpenImg: OpenImages. GT-k. Loc: GT-known Loc.}
        \label{table:hyper_select}
\renewcommand{\arraystretch}{1}
\renewcommand{\tabcolsep}{1.5pt}
\begin{tabular}{l|c|c|c|c|c}
\Xhline{2.\arrayrulewidth}
\hline 
\multirow{2}{*}{\textbf{Hyperparameters}} & \multicolumn{2}{c|}{\textbf{GT-k. Loc}} & \textbf{PxAP} & \multicolumn{2}{c}{\textbf{mIoU}} \\
\cline{2-6}
& \textbf{\texttt{CUB-200}} & \textbf{\texttt{ILSVRC}} & \textbf{\texttt{OpenImg}} & \textbf{\texttt{PASCAL}} & \textbf{\texttt{COCO}}\\
\Xhline{2.\arrayrulewidth}
\hline 
$\alpha$=0.2, $\beta$=1.2, $\lambda$=1.0 &95.29 & 71.75 & 66.67 & \textbf{57.73} & 36.79\\   

$\alpha$=0.5, $\beta$=1.5, $\lambda$=1.0 &\textbf{95.41} & 71.87 & \textbf{66.86} & 57.68 & {\textbf{36.91}}\\   

$\alpha$=0.7, $\beta$=1.7, $\lambda$=1.0 &95.35 & 71.94 & 66.74 & 57.55 & 36.84\\   

$\alpha$=1.0, $\beta$=2.0, $\lambda$=1.0 &95.26 & \textbf{72.00} & 66.52 & 57.41 & 36.67\\   

$\alpha$=1.5, $\beta$=2.5, $\lambda$=1.0 &95.08 & 71.89 & 66.39 & 57.16 & 36.43\\   
\hline 
Report &\textbf{95.41} & \textbf{72.00} & \textbf{66.86} & \textbf{57.73} & \textbf{36.91}\\   
\hline 
\Xhline{2.\arrayrulewidth}
\end{tabular}
\end{table}

\begin{table}[t]
    \small
\centering 
\caption{\textbf{The mIoU results of inserting the generator after different layers} with ResNet50 backbone.}
\label{table:layer_WSSS}
\renewcommand{\arraystretch}{1}
\renewcommand{\tabcolsep}{4pt}
\begin{tabular}{l|c|c|c|c}
\Xhline{2.\arrayrulewidth}
\hline 
\multirow{2}{*}{\textbf{Dataset}} & \multicolumn{4}{c}{\textbf{Different layers}}  \\
\cline{2-5}
& \textbf{Layer 1} & \textbf{Layer 2} & \textbf{Layer 3} & \textbf{Layer 4}\\
\Xhline{2.\arrayrulewidth}
\hline 
\textbf{\texttt{PASCAL VOC 2012}} &28.6 & 41.2 & \textbf{57.7} &55.3\\   
\hline 
\textbf{\texttt{MS COCO 2014}} &17.2 & 25.3 & \textbf{36.9} & 34.8\\   
\hline  
\Xhline{2.\arrayrulewidth}
\end{tabular}
\end{table}

Although three hyperparameters are included in the total loss, in practice, we simply follow a principle of $\beta = \alpha +\lambda$, so that $\mathcal{L}_{ac}$ is balanced with the losses $\mathcal{L}_{frg}$ and $\mathcal{L}_{bas}$. Meanwhile, in finding the most suitable ratio between $\alpha$ and $\lambda$, for simplicity, $\lambda$ is fixed at 1 on both WSOL and WSSS. Therefore, when $\alpha$ is determined, $\beta$ and $\lambda$ are also determined. In Table~\ref{table:hyper_select}, we provide the results of different combinations of hyperparameters on five datasets. When the $\alpha$ changes from 0.2 to 1.5, the effect on the results is limited with less than 0.6\% change. In fact, following the settings of $\alpha=0.5,\beta=1.5,\lambda=1.0$ is feasible for all datasets, with very little change compared to the results reported in the paper. The above experiments illustrate that it is easy to find a suitable set of hyperparameters on different datasets.

\begin{table}[t]
    \small
\centering
\caption{\textbf{Effects of applying BAS on different baseline methods}, including mIoU of the initial seed (\textbf{Seed}) and the pseudo ground-truth mask (\textbf{Mask}) on the \texttt{PASCAL VOC 2012} training set.}
        \label{table:voc_acc_train}
\renewcommand{\arraystretch}{1}
\renewcommand{\tabcolsep}{4pt}
\begin{tabular}{l|c|c|c}
\Xhline{2.\arrayrulewidth}
\hline 
\textbf{Method} & \textbf{Venue} & \textbf{Seed} & \textbf{Mask} \\
\Xhline{2.\arrayrulewidth}
\hline
IRN~\citep{ahn2019weakly} & CVPR$19$ & 48.8  & 66.3\\
SC-CAM~\citep{chang2020weakly} & CVPR$20$ & 50.9 & 63.4\\
SEAM~\citep{wang2020self} & CVPR$20$ & 55.4 & 63.6\\
CONTA~\citep{zhang2020causal} & NeurIPS$20$ & 48.8 & 67.9\\
CDA~\citep{su2021context} & ICCV$21$ & 50.8 & 67.7 \\
CSE~\citep{kweon2021unlocking} & ICCV$21$ & 56.0 & $-$\\
RIB~\citep{lee2021reducing} & NeurIPS$21$ & 56.5 & 68.6\\
ReCAM~\citep{chen2022class} & CVPR$22$ & 54.8 & 70.9\\
CLIMS~\citep{xie2022clims} & CVPR$22$ & 56.6 & 70.5\\
AdvCAM~\citep{lee2022anti} & TPAMI$22$ & 55.6 & 69.9\\

\hline
\rowcolor{mygray}
\textbf{Ours} & This Work & \textbf{57.7} & $-$\\

\rowcolor{mygray}
\textbf{Ours + IRN} & This Work & \textbf{58.2} & \textbf{71.1}\\

\hline
AdvCAM + CDA & $-$ & 55.5 & 69.3\\
ReCAM + CDA & $-$ & 54.5 & 70.5 \\
\rowcolor{mygray} 
\textbf{Ours + CDA} & This Work & \textbf{58.8} & \textbf{71.0}\\

\hline 
CDA + AdvCAM & $-$ & 55.5 & 69.3\\
ReCAM + AdvCAM & $-$ &56.6 & 70.9\\
\rowcolor{mygray} 
\textbf{Ours + AdvCAM} & This Work & \textbf{59.8} & \textbf{71.5}\\

\hline
\Xhline{2.\arrayrulewidth}
\end{tabular}
\end{table}

\begin{table*}[t]
\renewcommand{\arraystretch}{1}
\renewcommand{\tabcolsep}{1.4pt}
\small
\centering
\caption{\textbf{Semantic segmentation performance gains for per-class} on \textbf{\texttt{PASCAL VOC 2012}}.}
\label{table:per class}
\begin{tabular}{l|ccccccccccccccccccccc|c}
\Xhline{2.\arrayrulewidth}
\hline 
\textbf{Method} & \rotatebox{90}{\textbf{bkg}} & \rotatebox{90}{\textbf{aero}} & \rotatebox{90}{\textbf{bike}} & \rotatebox{90}{\textbf{bird}} & \rotatebox{90}{\textbf{boat}} & \rotatebox{90}{\textbf{bottle}} & \rotatebox{90}{\textbf{bus}} & \rotatebox{90}{\textbf{car}} & \rotatebox{90}{\textbf{cat}} & \rotatebox{90}{\textbf{chair}} & \rotatebox{90}{\textbf{cow}} & \rotatebox{90}{\textbf{table}} & \rotatebox{90}{\textbf{dog}} & \rotatebox{90}{\textbf{horse}} & \rotatebox{90}{\textbf{mbk}} & \rotatebox{90}{\textbf{person}} & \rotatebox{90}{\textbf{plant}} & \rotatebox{90}{\textbf{sheep}} & \rotatebox{90}{\textbf{sofa}} & \rotatebox{90}{\textbf{train}} & \rotatebox{90}{\textbf{tv}} & \textbf{mIoU} \\
\Xhline{2.\arrayrulewidth}
\hline 
IRN & 80.0 & 45.8 & 30.1 & 41.5 & \textbf{38.8} &45.6 & 61.4 & 52.6 & 43.3 & \textbf{29.0} & 56.8 & 40.7 & 44.2 & 53.3 & 62.7 & 51.1 & 45.2 & 63.3 & 44.6 & 50.5 & \textbf{44.2} & 48.8 \\
\rowcolor{mygray}
\textbf{\enspace+ Ours} & \textbf{82.3} & \textbf{50.1} & \textbf{39.1} & \textbf{54.9} & 30.4 & \textbf{56.4} & \textbf{76.7} & \textbf{58.5} & \textbf{73.5} & 27.6 & \textbf{72.7} & \textbf{50.3} & \textbf{68.9} & \textbf{70.3} & \textbf{73.7} & \textbf{59.7} & \textbf{48.7} & \textbf{79.2} & \textbf{52.0} & \textbf{55.1} & 42.1 & \textbf{58.2} \\
\hline
CDA & 80.2 & 44.6 & 28.9 & 45.8 & \textbf{36.9} & 52.9 & 65.2 & 54.4 & 55.7 & \textbf{28.4} & 57.5 & 42.1 & 54.1 & 54.2 & 60.2 & 54.3 & 47.7 & 65.7 & 46.6 & 49.3 & \textbf{43.9} & 50.8\\
\rowcolor{mygray}
\textbf{\enspace+ Ours} & \textbf{82.6} & \textbf{50.1} & \textbf{39.0} & \textbf{56.1} & 30.6 & \textbf{57.5} & \textbf{77.1} & \textbf{59.8} & \textbf{75.3} & 27.6 & \textbf{73.8} & \textbf{49.6} & \textbf{70.5} & \textbf{71.4} & \textbf{73.4} & \textbf{60.1} & \textbf{51.3} & \textbf{80.6} & \textbf{51.1} & \textbf{54.6} & 41.4 & \textbf{58.8} \\
\hline
AdvCAM & 81.3 & 50.6 & 33.5 & 57.3 & \textbf{36.9} & 53.1 & 67.9 & 54.8 & 64.8 & \textbf{35.0} & 68.4 & 42.0 & 58.4 & 67.9 & 67.1 & 56.0 & 42.6 & 76.1 & 48.5 & 56.8 & \textbf{45.9} & 55.5\\
\rowcolor{mygray}
\textbf{\enspace+ Ours} & \textbf{83.1} & \textbf{52.1} & \textbf{39.4} & \textbf{59.8} & 31.4 & \textbf{58.3} & \textbf{77.5} & \textbf{60.2} & \textbf{75.9} & 30.3 & \textbf{76.2} & \textbf{49.5} & \textbf{70.4} & \textbf{73.5} & \textbf{74.8} & \textbf{62.3} & \textbf{48.3} & \textbf{82.0} & \textbf{51.6} & \textbf{56.9} & 43.0 & \textbf{59.8} \\

\hline
\Xhline{2.\arrayrulewidth}
\end{tabular}
\end{table*}

\begin{table}[t]
    \small
\centering
\caption{\textbf{Performance comparison of WSSS methods} in terms of mIoU (\%) on the \texttt{PASCAL VOC 2012} val and test sets. Sup.: supervision. $\mathcal{F}$: full supervision. $\mathcal{I}$: image-level supervision. $\mathcal{S}$: saliency map supervision.}
        \label{table:voc_acc}
    \renewcommand{\arraystretch}{1}
\renewcommand{\tabcolsep}{1pt}
\begin{tabular}{l|c|c|c|c}
\Xhline{2.\arrayrulewidth}
\hline 
\textbf{Method} & \textbf{Venue} & \textbf{Sup.} &  \textbf{Val} & \textbf{Test} \\
\Xhline{2.\arrayrulewidth}
\hline 
\multicolumn{4}{l}{\textbf{Full supervision.}} \\
DeepLabV2~\citep{chen2017deeplab} & TPAMI$18$ & $\mathcal{F}$ & $77.6$ &$79.7$ \\
WideResNet38~\citep{wu2019wider} & PR$19$ & $\mathcal{F}$ & $80.8$ &$82.5$ \\
\Xhline{2.\arrayrulewidth}
\hline  
\multicolumn{4}{l}{\textbf{Image-level supervision + Saliency maps.}} \\
OAA~\citep{jiang2021online} & TPAMI$21$ & $\mathcal{I}$ + $\mathcal{S}$ & 66.1 & 67.2 \\
AuxSegNet~\citep{xu2021leveraging} & ICCV$21$ & $\mathcal{I}$ + $\mathcal{S}$ & 69.0 & 68.6 \\
AdvCAM~\citep{lee2022anti} & TPAMI$22$ & $\mathcal{I}$ + $\mathcal{S}$ &  71.3 &  71.2 \\

\Xhline{2.\arrayrulewidth}
\hline 
\multicolumn{4}{l}{\textbf{Image-level supervision only.}} \\
IRN~\citep{ahn2019weakly} & CVPR$19$ & $\mathcal{I}$  & 63.5 & 64.8 \\
BES~\citep{chen2020weakly} & ECCV$20$ & $\mathcal{I}$ & 65.7 & 66.6 \\
CONTA~\citep{zhang2020causal} & NeurIPS$20$ & $\mathcal{I}$ & 65.3 & 66.1\\
IAL~\citep{wang2020weakly} & IJCV$20$ & $\mathcal{I}$ &  62.0 & 62.4 \\
ADL~\citep{choe2020attention} & TPAMI$20$ & $\mathcal{I}$ & 53.7 & 54.7 \\
LIID~\citep{liu2020leveraging} & TPAMI$20$ & $\mathcal{I}$  & 66.5 & 67.5\\
RIB~\citep{lee2021reducing} & NeurIPS$21$ & $\mathcal{I}$  & 68.3 & 68.6 \\
CDA~\citep{su2021context} & ICCV$21$ & $\mathcal{I}$ & 65.8 & 66.4 \\
ECS~\citep{sun2021ecs} & ICCV$21$ & $\mathcal{I}$ &  66.6 & 67.6 \\
PMM~\citep{li2021pseudo} & ICCV$21$ & $\mathcal{I}$ & 68.5 & 69.0 \\
CSE~\citep{kweon2021unlocking} & ICCV$21$ & $\mathcal{I}$ & 68.4 & 68.2 \\
CPN~\citep{zhang2021complementary} & ICCV$21$ & $\mathcal{I}$ & 67.8 & 68.5\\
{A}$^{2}$GNN~\citep{zhang2021affinity} & TPAMI$21$ & $\mathcal{I}$ &  66.8 & 67.4\\
AFA~\citep{ru2022learning} & CVPR$22$ & $\mathcal{I}$  & 66.0 & 66.3\\
Du et al.~\citep{du2022weakly} & CVPR$22$ & $\mathcal{I}$  & 67.7 & 67.4\\
ReCAM~\citep{chen2022class} & CVPR$22$ & $\mathcal{I}$  & 68.5 & 68.4\\
SIPE~\citep{chen2022self} & CVPR$22$ & $\mathcal{I}$  & 68.8 & 69.7\\
MCIS~\citep{wang2022looking} & TPAMI$22$ & $\mathcal{I}$ &  66.2 & 66.9 \\
AdvCAM~\citep{lee2022anti} & TPAMI$22$ & $\mathcal{I}$ &  68.1 & 68.0 \\
\hline
\rowcolor{mygray}
\textbf{Ours} & This Work & $\mathcal{I}$ & \textbf{69.6} & \textbf{69.9} \\
\hline
\Xhline{2.\arrayrulewidth}
\end{tabular}
\end{table}

\myPara{Generator after different layers.} In Table~\ref{table:layer_WSSS}, we report the mIoU results of inserting the generator after different layers of ResNet50. Since the generator contains only one convolution layer, the semantic representation of the generated localization map depends mainly on the reused backbone part. Therefore, inserting the generator after \textbf{layer 1} or \textbf{layer 2} will result in insufficient semantic representation and poor segmentation performance, as presented in Table~\ref{table:layer_WSSS}. In addition, inserting the generator after \textbf{4} does not perform better than \textbf{layer 3}, reducing 2.4\% and 2.1\% mIoU on \texttt{PASCAL VOC 2012} and  \texttt{MS COCO 2014}, respectively. It is mainly because the feature maps of \textbf{layer 4} are usually coarser than the feature maps of \textbf{layer 3}, hindering the acquisition of fine segmentation results.

\begin{figure*}[t]
\centering
	\begin{overpic}[width=1.\linewidth]{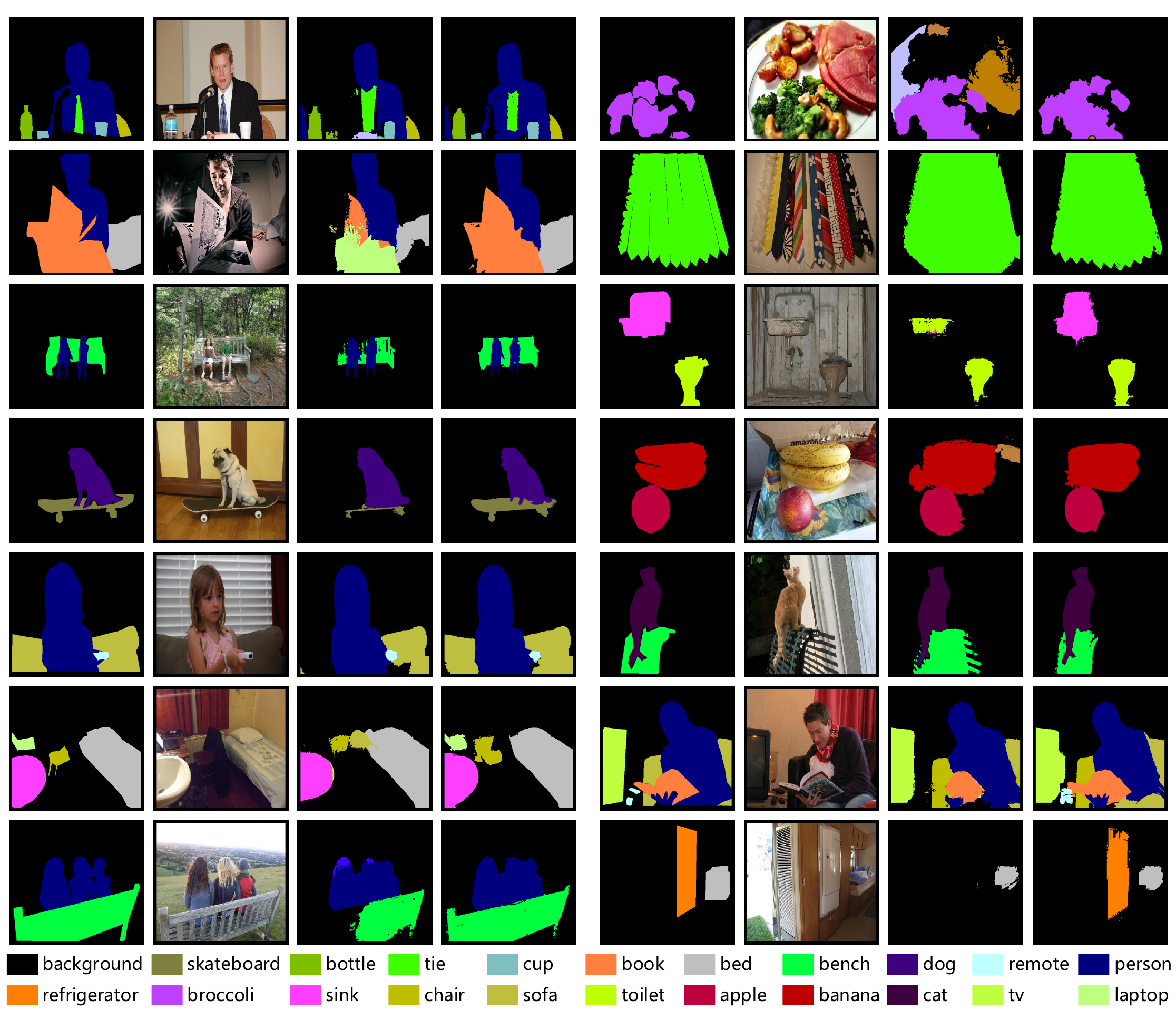}
	    \put(5, 84.8){{\small\textbf{GT}}}
	    \put(16, 84.8){{\small\textbf{Image}}}
	    \put(29., 84.8){{\small\textbf{IRN}}}
	    \put(41, 84.8){{\small\textbf{BAS}}}
	    
	    \put(55.0, 84.8){{\small\textbf{GT}}}
	    \put(66.8, 84.8){{\small\textbf{Image}}}
	    \put(79.8, 84.8){{\small\textbf{IRN}}}
	    \put(91.8, 84.8){{\small\textbf{BAS}}}
	    
\end{overpic}
\caption{\textbf{Examples of semantic segmentation results} on \texttt{MS COCO 2014} for IRN and BAS (with IRN).}
\label{mask_coco}
\end{figure*}

\subsection{Results on \texttt{PASCAL VOC 2012} Dataset \label{section:result_pascal}}

\myPara{Quality of Initial Seed and Pseudo Labels.} Table~\ref{table:voc_acc_train} compares the quality of the initial seed and the pseudo ground-truth masks on the \texttt{PASCAL VOC 2012} training set. For the initial seed, we achieve a mIoU of \textbf{57.7\%}, exceeding the previous method by a large margin. Compared with the state-of-the-art method CLIMS~\citep{xie2022clims}, which uses both ResNet50 and CLIP~\citep{radford2021learning} networks in the seed generation phase, while BAS uses only ResNet50 network and achieves a gain of \textbf{1.1\%}. Further, after normalizing the seeds generated by our method and by other methods and adding them together, BAS can combine with various baseline methods and significantly improve their segmentation quality by providing high quality foreground prediction maps. As shown in Table~\ref{table:voc_acc_train}, the proposed BAS improves the IRN~\citep{ahn2019weakly} by \textbf{9.4\%} mIoU, which is a remarkable boost. In addition, we achieve the best results with \textbf{59.8\%} mIoU when applying BAS to AdvCAM~\citep{lee2022anti}. We also add the initial seeds of the different methods for a fair comparison in Table~\ref{table:voc_acc_train}. It is obvious that combining with BAS brings more remarkable improvement than combining with other methods. This is because BAS can produce high and balanced responses on the object, which benefits other methods significantly. We report the per-class mean IoU in Table~\ref{table:per class}. Although our method achieves consistent improvement on the above baseline methods, it does not perform well in some categories. This is because the classification network has difficulty distinguishing between objects and class-related contexts, especially in some categories, e.g., boats and water, TV and programs on TV, which in turn limits the localization ability of BAS. Fig.~\ref{seed_voc} shows the visual comparison of the initial seed generated by BAS and IRN. It can be clearly noticed that our method has better performance in capturing the whole object area with a high confidence score. For the pseudo ground-truth mask, after refinement by IRN~\citep{ahn2019weakly}, we achieve \textbf{4.8\%}, \textbf{3.3\%}, and \textbf{1.6\%} gains when BAS is deployed on IRN, CDA, and AdvCAM, respectively, which illustrates the effectiveness of the proposed method. BAS allows to obtain a better foreground-background segmentation and thus provides a strong support for the seed generation stage of the WSSS task.

\myPara{Quality of Segmentation.} To further validate the effectiveness of our method, we employ the pseudo segmentation labels to directly train a semantic segmentation network. Table~\ref{table:voc_acc} presents the segmentation results of the proposed BAS (with IRN) and other methods on the \texttt{PASCAL VOC 2012} dataset. It is observed that our BAS exceeds previous methods under the same level of supervision, with \textbf{69.6\%} and \textbf{69.9\%} mIoU on the val and test sets. Compared to the latest method ReCAM~\citep{chen2022class}, with the same backbone network, we achieve a \textbf{1.1\%} mIoU improvement on val set and \textbf{1.5\%} on test set. We also show some qualitative segmentation results in Fig.~\ref{mask_voc}. Compared with IRN, BAS demonstrates more robustness to various challenging scenarios, such as various sized objects, complex environments, and multi-instance situations.

\begin{table}[t]
    \small
\centering
    \renewcommand{\arraystretch}{1}
\renewcommand{\tabcolsep}{1pt}
\caption{\textbf{Evaluation results on \texttt{MS COCO 2014} validation set}. Sal: Saliency. Bac: Backbone. WR38: WideResNet38. R50/101: ResNet50/101.}
        \label{table:coco_val}
\begin{tabular}{l|c|c|c|c}
\Xhline{2.\arrayrulewidth}
\hline 
\textbf{Method} & \textbf{Venue} & \textbf{Bac.} & \textbf{Sal.} & \textbf{mIoU}  \\
\Xhline{2.\arrayrulewidth}
\hline 
IRN~\citep{ahn2019weakly} & CVPR$19$ & R101 &  &  41.4 \\
IAL~\citep{wang2020weakly} & IJCV$20$ & VGG16 &  & 27.7 \\
ADL~\citep{choe2020attention} & TPAMI$20$ & VGG16 &  \checkmark & 30.8 \\
CONTA~\citep{zhang2020causal} & NeurIPS$20$ & R50 &  & 33.4 \\
EPS~\citep{lee2021railroad} & ICCV$21$ & WR38 & \checkmark & 35.7 \\
CSE~\citep{kweon2021unlocking} & ICCV$21$ & WR38 &  & 36.4 \\
PMM~\citep{li2021pseudo} & ICCV$21$ & WR38 &  & 36.7 \\
RIB~\citep{lee2021reducing} & NeurIPS$21$ & R101 &  & 43.8 \\
ReCAM~\citep{chen2022class} & CVPR$22$ & R50 &  & 44.1 \\
L2G~\citep{jiang2022l2g} & CVPR$22$ & R101 & \checkmark & 44.2 \\
AdvCAM~\citep{lee2022anti} & TPAMI$22$ & R101 &  & 44.4 \\
\hline
\rowcolor{mygray}
\textbf{Ours} & This Work & R50 &  & \textbf{45.1} \\
\hline
\Xhline{2.\arrayrulewidth}
\end{tabular}
\end{table}

\subsection{Results on \texttt{MS COCO 2014} Dataset}

The accuracy of the proposed method and other state-of-the-art approaches on the \texttt{MS COCO 2014} validation set is compared in Table~\ref{table:coco_val}. Our BAS based on IRN achieves a mIoU value of \textbf{45.1\%}, exceeding all previous methods. Compared to the previous best model AdvCAM (ResNet101 is adopted as the backbone network), we use a smaller ResNet50 as the backbone, but achieve better results. In particular, we surpass our baseline method IRN~\citep{ahn2019weakly} by \textbf{3.7\%} mIoU. Fig.~\ref{seed_coco} demonstrates the visual comparison of the initial seed obtained by CAM~\citep{zhou2016learning} and our method. Qualitative experiments show that the proposed BAS can capture more object areas compared to CAM, especially for large objects and multiple instances. In addition, BAS can achieve balanced and comprehensive responses on the target regions across various categories. Fig.~\ref{mask_coco} shows some examples of semantic segmentation masks on \texttt{MS COCO 2014} produced by IRN and by BAS (with IRN). It is observed that our method employed on the IRN can achieve more accurate segmentation and show a better demarcation between different objects, because the proposed BAS can provide a more complete and accurate seed region compared to IRN.

\begin{figure}[t]
\small
    \centering
    \begin{overpic}[width=0.98\linewidth]{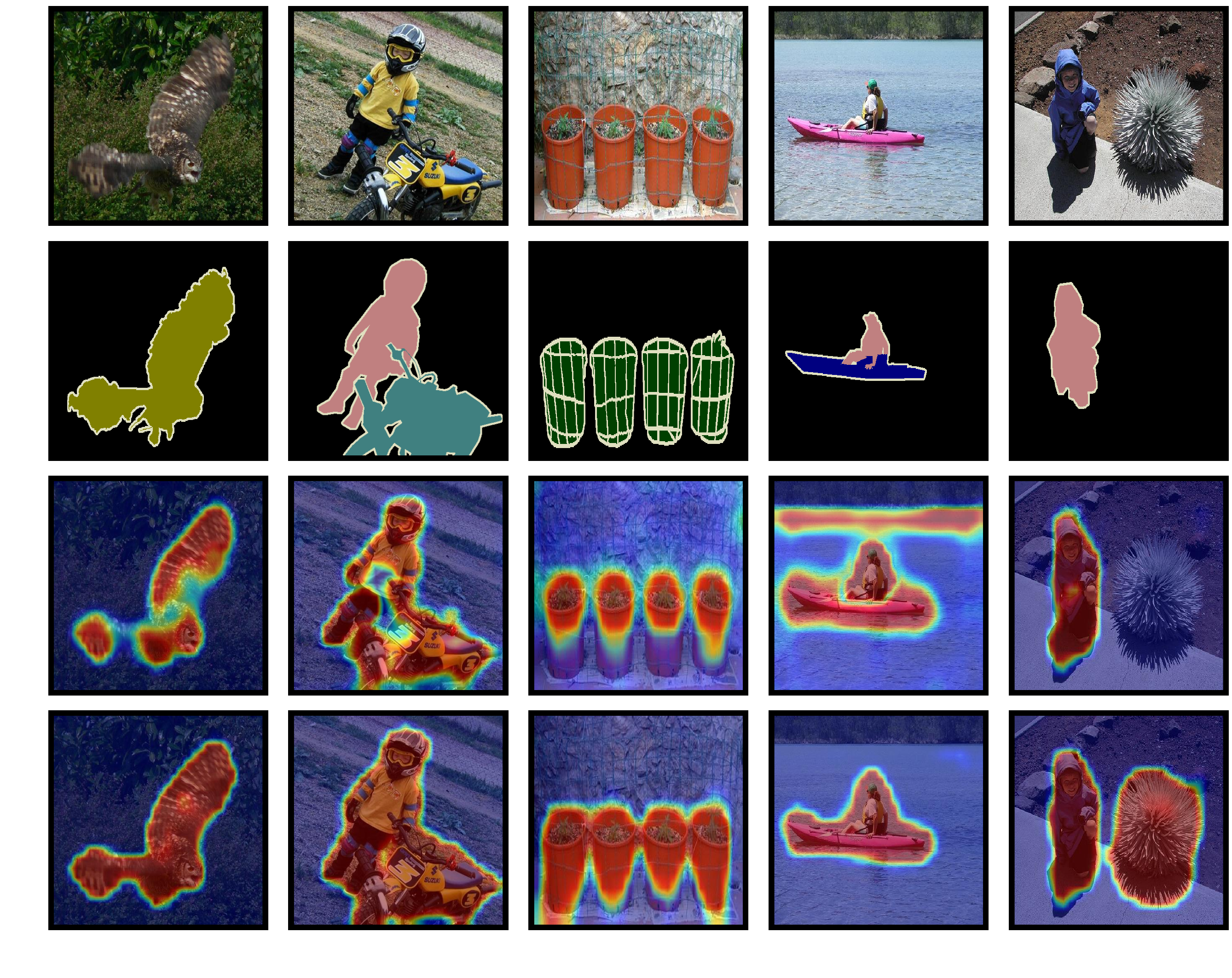}
    \put(10.5, 0.5){\small\textbf{(a)}}
    \put(29.5, 0.5){\small\textbf{(b)}}
    \put(49.5, 0.5){\small\textbf{(c)}}
    \put(68.8, 0.5){\small\textbf{(d)}}
    \put(88.8, 0.5){\small\textbf{(e)}}
    \put(0.3, 5.){\rotatebox{90}{\small\textbf{Agnostic}}}
    \put(0.3, 25.){\rotatebox{90}{\small\textbf{Specific}}}
    \put(0.3, 49.){\rotatebox{90}{\small\textbf{GT}}}
    \put(0.3, 65.){\rotatebox{90}{\small\textbf{Image}}}
    \end{overpic}
    
    \caption{\textbf{Class-agnostic foreground map \bm{$vs$} class-specific  localization map} in the following five aspects:  (a) Completeness. (b) Connectivity. (c) Less noise. (d) Identifying class-related background. (e) Class-aware.}
    \label{fig:foreground}
\end{figure}
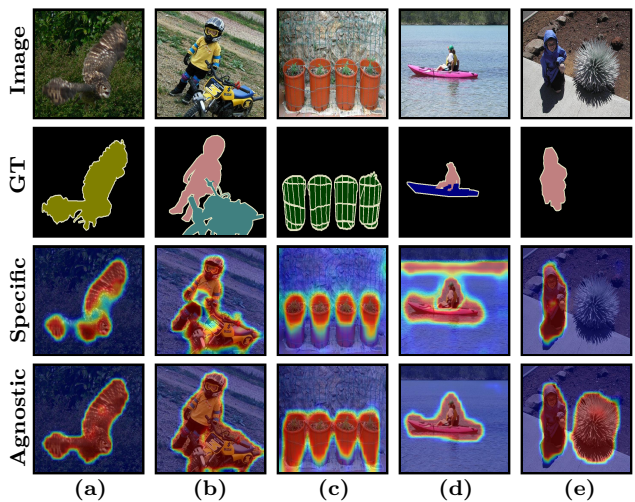

\subsection{Analysis \label{sec:analysis}}

In this section, we will explore how to fully leverage BAS, especially focusing on its crucial background activation suppression loss. Furthermore, we aim to enhance the segmentation capability of BAS by integrating it with other methods.

\myPara{Class-agnostic foreground map.} Different from the CAM-based approaches to extract class activation maps from the classifier, the proposed BAS obtains localization maps through an extra generator. In addition to generating class-specific localization maps, BAS can also produce a class-agnostic foreground map by providing suitable objective functions. To this end, we consider all the classes existing in the image as a foreground class and sum the $\mathcal{L}_{bas}$ of existing classes to supervise the foreground map. In this way, the foreground map can be fully trained from the entire dataset. As shown in Fig.~\ref{fig:foreground}, the class-agnostic foreground map localizes objects more completely and robustly than the class-specific localization map, and generates less noise. However, the foreground map is unable to distinguish objects of different categories and often identifies objects that are not in the target classes as shown in Fig.~\ref{fig:foreground} (e). To utilize the foreground map to improve the performance of class-specific localization maps, we follow an intuitive idea that the foreground map usually covers all class-specific localization maps and has higher segmentation quality. If the class-specific localization map has a higher response in some regions than the foreground map, it may be caused by noise or confusing background, as shown in Fig.~\ref{fig:foreground} (c) and Fig.~\ref{fig:foreground} (d). Therefore, we should weaken the response in these regions by directly replacing them with the response in the foreground map or averaging the response of both maps. The experimental results in Table~\ref{table:foreground} show that both strategies can improve the quality of the initial seed and hence increase the accuracy of the pseudo ground-truth mask. The best results are achieved by the average approach which not only reduces the response of the uncertain region but also combines the prediction probabilities of both class-agnostic and class-specific maps. It improves the initial seed and pseudo ground-truth mask by \textbf{0.6\%} and \textbf{0.5\%} mIoU results.

\begin{table}[t]
    \small
\centering 
\caption{\textbf{Applying the class-agnostic foreground map to the class-specific localization maps} with different strategies on the \texttt{PASCAL VOC 2012}.}
        \label{table:foreground}
\renewcommand{\arraystretch}{1}
\renewcommand{\tabcolsep}{4.5pt}
\begin{tabular}{l|c|c|c|c|c}
\Xhline{2.\arrayrulewidth}
\hline 
\textbf{Method} & \textbf{Strategy} &\textbf{Seed} & \textbf{Mask} & \textbf{Val} & \textbf{Test} \\ 
\Xhline{2.\arrayrulewidth}
\hline 
\textbf{Ours} & $-$ & 57.7 &71.1&69.6&69.9\\ 
\rowcolor{mygray}
\hline
\textbf{w/ Foreground} & Replace & 57.9 &71.4& 69.9& 70.0\\  
\rowcolor{mygray}
\hline
\textbf{w/ Foreground} &  Average  & \textbf{58.3} &\textbf{71.6}& \textbf{70.3} & \textbf{70.1}\\  
\hline
\Xhline{2.\arrayrulewidth}
\end{tabular}
\end{table}

\myPara{Improve the quality of BAS.} As analyzed in Section~\ref{section:result_pascal}, It can be noted that BAS does not perform well in some categories, which is usually due to the co-occurring context providing support to the classification discrimination, causing the localization map to learn the context. To alleviate this problem, we apply the proposed BAS to the W-OoD~\citep{lee2022weakly} method, which uses additional out-of-distribution data to address the spurious relevance of the background, such as boat-water and aeroplane-sky/runway. As presented in Table~\ref{table:ood}, benefiting from the strong discriminative ability of the classification network in W-OoD method, BAS can achieve better performance, with a \textbf{1.8\%} mIoU improvement on the initial seed, including \textbf{16.0\%} and \textbf{7.1\%} mIoU gains on the boat and aeroplane categories, respectively. After applying IRN and DeepLabV2, BAS w/ W-OoD obtains \textbf{71.3\%} and \textbf{71.1\%} mIoU on \texttt{PASCAL VOC 2012} val and test sets. In addition, we apply CLIMS~\citep{xie2022clims} to BAS to suppress the co-occurring background by using natural language supervision in CLIP~\citep{radford2021learning}, which also significantly improves the quality of the initial seed and brings a \textbf{8.9\%} boost in the boat category. Consequently, BAS w/ CLIMS obtains \textbf{70.6\%} and \textbf{70.9\%} mIoU on the val set and test set, substantially improving the segmentation ability of BAS.

\begin{table}[t]
    \small
\centering 
\caption{\textbf{Applying BAS to CLISM and W-OoD} on the \texttt{PASCAL VOC 2012}.}
        \label{table:ood}
\renewcommand{\arraystretch}{1}
\renewcommand{\tabcolsep}{3pt}
\begin{tabular}{l|c|c|c|c}
\Xhline{2.\arrayrulewidth}
\hline 
\textbf{Method} & \textbf{Seed} & \textbf{Mask} & \textbf{Val} & \textbf{Test}\\
\Xhline{2.\arrayrulewidth}
\hline 
\textbf{Ours}&57.7 & 71.1 & 69.6 & 69.9\\ 
\rowcolor{mygray}
\hline
\textbf{w/ CLIMS~\citep{xie2022clims}} &   \textbf{59.0} & \textbf{72.3} & \textbf{70.6} & \textbf{70.9}\\  
\rowcolor{mygray}
\hline
\textbf{w/ W-OoD~\citep{lee2022weakly}} &   \textbf{59.5} &\textbf{72.7}& \textbf{71.3} & \textbf{71.1}\\  
\hline
\Xhline{2.\arrayrulewidth}
\end{tabular}
\end{table}

\begin{figure*}[t]
    \centering
    \begin{overpic}[width=1.\linewidth]{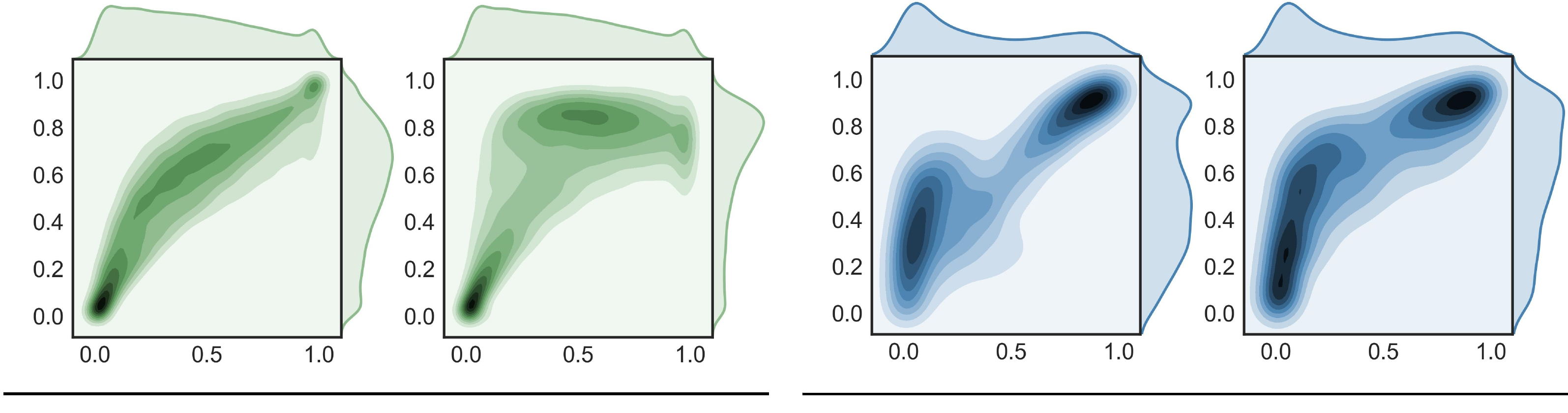}
	    \put(8.5, 1.8){\small\textbf{Object Size}}
	    \put(32., 1.8){\small\textbf{Object Size}}
	    \put(60., 1.8){\small\textbf{Object Size}}
	    \put(84., 1.8){\small\textbf{Object Size}}
	    
	    \put(16.2 , 5.5){\small\textbf{CAM}}
	    \put(40.5, 5.5){\small\textbf{BAS}}
	    \put(68.3, 5.5){\small\textbf{IRN}}
	    \put(91.5, 5.5){\small\textbf{BAS}}
	    
	    \put(0, 12.){\rotatebox{90}{\small\textbf{IoU}}}
	    \put(51, 12.){\rotatebox{90}{\small\textbf{IoU}}}
	    \put(22., -0.3){\small\textbf{\texttt{ILSVRC}}}
	    \put(68., -0.3){\small\textbf{\texttt{PASCAL VOC 2012}}}
    \end{overpic}
    \caption{\textbf{Limitation.} The density distribution map about IoU and object size. For WSOL, the experiment is implemented on \texttt{ILSVRC} and bounding boxes are used to measure IoU and object size. For WSSS, experimental results are calculated by pixel-level masks on the \texttt{PASCAL VOC 2012} training set at the seed phase.}
    \label{limitation}
\end{figure*}

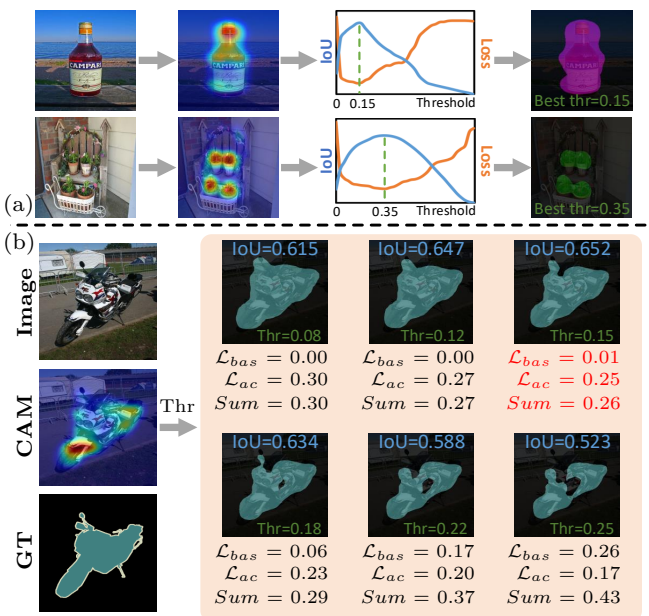
\begin{figure}[t]
    \centering
    \begin{overpic}[width=1.\linewidth]{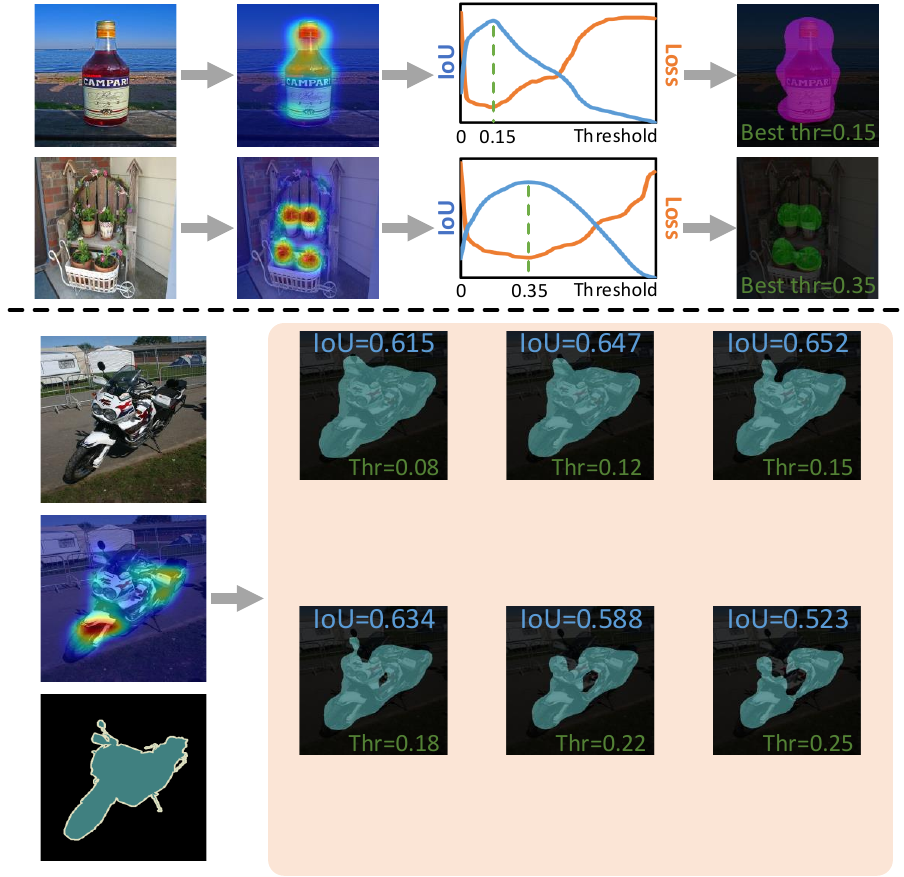}
	    \put(1., 8.3){\rotatebox{90}{\small\textbf{GT}}}
     \put(1., 26.){\rotatebox{90}{\small\textbf{CAM}}}
     \put(1., 45.){\rotatebox{90}{\small\textbf{Image}}} 

     \put(23.3, 33.3){\scriptsize{Thr}}

     \put(-1.2, 65.){\small{(a)}}
     \put(-1.2, 59){\small{(b)}}
     
     \put(32.0, 41.){\scriptsize{$\mathcal{L}_{bas}$ = 0.00}}
     \put(33.4, 37.5){\scriptsize{$\mathcal{L}_{ac}$ = 0.30}}
     \put(31.8, 33.7){\scriptsize{$Sum$ = 0.30}}
 
     \put(55.0, 41){\scriptsize{$\mathcal{L}_{bas}$ = 0.00}}
     \put(56.4, 37.5){\scriptsize{$\mathcal{L}_{ac}$ = 0.27}}
     \put(54.8, 33.7){\scriptsize{$Sum$ = 0.27}}

     \put(77.8, 41){\textcolor{wpy_1}{\scriptsize{$\mathcal{L}_{bas}$ = 0.01}}}
     \put(79.2, 37.5){\textcolor{wpy_1}{\scriptsize{$\mathcal{L}_{ac}$ = 0.25}}}
     \put(77.6, 33.7){\textcolor{wpy_1}{\scriptsize{$Sum$ = 0.26}}}

     \put(32.0, 10.3){\scriptsize{$\mathcal{L}_{bas}$ = 0.06}}
     \put(33.4, 6.9){\scriptsize{$\mathcal{L}_{ac}$ = 0.23}}
     \put(31.8, 3.0){\scriptsize{$Sum$ = 0.29}}

     \put(55.0, 10.3){\scriptsize{$\mathcal{L}_{bas}$ = 0.17}}
     \put(56.4, 6.9){\scriptsize{$\mathcal{L}_{ac}$ = 0.20}}
     \put(54.8, 3.0){\scriptsize{$Sum$ = 0.37}}

     \put(77.8, 10.3){\scriptsize{$\mathcal{L}_{bas}$ = 0.26}}
     \put(79.2, 6.9){\scriptsize{$\mathcal{L}_{ac}$ = 0.17}}
     \put(77.6, 3.0){\scriptsize{$Sum$ = 0.43}}
     
    \end{overpic}
    
    \caption{\textbf{Finding image-specific threshold by BAS.} (a) IoU-threshold curve and Loss-threshold curve. Loss indicates the summation of $\mathcal{L}_{bas}$ and $\mathcal{L}_{ac}$. (b) Process of finding the image-specific threshold by using $\mathcal{L}_{bas}$ and $\mathcal{L}_{ac}$ as evaluation.}
    \label{fig:evaluate}
\end{figure}

\myPara{Finding image-specific threshold by BAS.} Unlike CAM, the proposed BAS designs a set of loss functions to evaluate the quality of the localization map and uses them for training, similarly, they are also suitable for the testing phase. As shown in Fig.~\ref{fig:evaluate} (a), it can be noted that the unbalanced response of CAM causes the segmentation performance heavily dependent on the threshold, while the optimal threshold value even varies significantly across images. It is obviously not appropriate to use a global threshold for the whole dataset. Therefore, we propose to find the image-specific threshold by employing background activation suppression loss $\mathcal{L}_{bas}$ and area constraint loss $\mathcal{L}_{ac}$ as the evaluation of the threshold values. As illustrated in Fig.~\ref{fig:evaluate} (b), we obtain a series of binary masks by changing the threshold values and input them into the AMC module to generate $\mathcal{L}_{bas}$ and $\mathcal{L}_{ac}$, the same process as in Fig.~\ref{fig:network}. Then, we simply add $\mathcal{L}_{bas}$ and $\mathcal{L}_{ac}$ together as the evaluation score and select the binary mask with the smallest evaluation score as the final result. Table~\ref{table:evaluate} compares the effect of this image-specific threshold post-processing with global threshold on different methods on the \texttt{PASCAL VOC 2012} training set. Experimental results show that the proposed post-processing approach is helpful to improve the segmentation quality by providing feedback on different thresholds to select the best threshold value specific to the image, especially for CAM~\citep{zhou2016learning} and CDA~\citep{su2021context}, bringing \textbf{0.7\%} and \textbf{0.5\%} mIoU improvement, respectively. However, the enhancement is limited when applying it to the proposed BAS, mainly because BAS produces few uncertainty regions and is not very sensitive to the threshold.


\begin{table}[t]
    \small
\centering 
\caption{\textbf{Effect of applying image-specific threshold} on different methods compared to global threshold on the \texttt{PASCAL VOC 2012} training set.}
        \label{table:evaluate}
\renewcommand{\arraystretch}{1}
\renewcommand{\tabcolsep}{2.5pt}
\begin{tabular}{l|c|c|c|c|c}
\Xhline{2.\arrayrulewidth}
\hline 
\multirow{2}{*}{\textbf{Threshold}} & \multicolumn{5}{c}{\textbf{Method}} \\
\cline{2-6}
& \textbf{CAM} & \textbf{CDA} & \textbf{ReCAM} & \textbf{AdvCAM} & \textbf{Ours}\\
\Xhline{2.\arrayrulewidth}
\hline 
\textbf{Global} &48.8 & 50.8 & 54.8 & 55.5 & 57.7\\   
\hline 
\rowcolor{mygray}
\textbf{Image-specific} &\textbf{49.5} & \textbf{51.3} & \textbf{55.0} & \textbf{55.7} & \textbf{57.8}\\   
\hline 
\Xhline{2.\arrayrulewidth}
\end{tabular}
\end{table}

\section{Discussion \label{sec:discussion}}

\myPara{Limitation.} In this section, we discuss the localization ability of BAS for different size objects. We first visualize the density distribution of the IoU about BAS and CAM~\citep{zhou2016learning} in Fig.~\ref{limitation}. It can be noted that BAS performs better on medium and large objects, but not enough on small objects. We believe the main reason is the following two aspects: the localization of small objects is an inherent problem of computer vision, on the other hand, the area constraint loss penalizes different size objects unequally and will penalize small objects less, which causes BAS cannot balance both large and small objects with only a single hyperparameter to adjust the area constraint loss.

\myPara{Future Works.}
In the future, there are two main aspects of work, \textbf{1)} improving the performance of the localization capability at different object sizes and \textbf{2)} further extending the application of BAS.

To solve the issue of inconsistent localization ability for different size objects, we would like to explore the following promising researches: \textbf{1)} The area constraint loss can be improved to allow different tolerance for objects of various sizes. \textbf{2)} Based on the fact that WSOL works better for localizing large objects, we can determine the approximate region of the objects in the first stage, and then crop and resize the corresponding region to convert the original small object into a larger one, thereby performing localization in the second stage. 

Apart from the above possible improvements, BAS can also be extended to weakly supervised instance segmentation (WSIS), since obtaining a high-quality locality map is also essential for WSIS.

\section{Conclusion \label{sec:conclusion}}
In this paper, we find previous FPM-based work using cross-entropy to facilitate the learning of foreground prediction maps, essentially by changing the activation value, and the activation value shows a higher correlation with the foreground mask. Thus, we propose a Background Activation Suppression (BAS) approach to promote the generation of foreground maps by an Activation Map Constraint (AMC) module, which facilitates the learning of foreground prediction maps mainly through the suppression of background activation. Extensive experiments on \texttt{CUB-200-2011} and \texttt{ILSVRC} verify the effectiveness of the proposed BAS, which surpasses previous methods by a large margin. In addition, BAS can also be extended on WSSS to enhance the seed quality of other methods by providing high quality foreground maps, and achieves the state-of-the-art performance on \texttt{PASCAL VOC 2012} and \texttt{MS COCO 2014}.

\newcommand{\tabincell}[2]{\begin{tabular}{@{}#1@{}}#2\end{tabular}}  


\normalem
\bibliographystyle{spbasic}      
{\bibliography{IJCV}}

\end{document}